\PassOptionsToPackage{unicode}{hyperref}
\PassOptionsToPackage{hyphens}{url}
\PassOptionsToPackage{dvipsnames,svgnames,x11names}{xcolor}
\documentclass[
]{article}
\usepackage{amsmath,amssymb}
\usepackage{iftex}
\ifPDFTeX
  \usepackage[T1]{fontenc}
  \usepackage[utf8]{inputenc}
  \usepackage{textcomp} 
\else 
  \usepackage{unicode-math} 
  \defaultfontfeatures{Scale=MatchLowercase}
  \defaultfontfeatures[\rmfamily]{Ligatures=TeX,Scale=1}
\fi
\usepackage{lmodern}
\ifPDFTeX\else
\fi
\IfFileExists{upquote.sty}{\usepackage{upquote}}{}
\IfFileExists{microtype.sty}{
  \usepackage[]{microtype}
  \UseMicrotypeSet[protrusion]{basicmath} 
}{}
\makeatletter
\@ifundefined{KOMAClassName}{
  \IfFileExists{parskip.sty}{%
    \usepackage{parskip}
  }{
    \setlength{\parindent}{0pt}
    \setlength{\parskip}{6pt plus 2pt minus 1pt}}
}{
  \KOMAoptions{parskip=half}}
\makeatother
\usepackage{xcolor}
\usepackage[margin=1in]{geometry}
\usepackage{graphicx}
\makeatletter
\def\maxwidth{\ifdim\Gin@nat@width>\linewidth\linewidth\else\Gin@nat@width\fi}
\def\maxheight{\ifdim\Gin@nat@height>\textheight\textheight\else\Gin@nat@height\fi}
\makeatother
\setkeys{Gin}{width=\maxwidth,height=\maxheight,keepaspectratio}
\makeatletter
\def\fps@figure{htbp}
\makeatother
\providecommand{\tightlist}{%
  \setlength{\itemsep}{0pt}\setlength{\parskip}{0pt}}
\newlength{\cslhangindent}
\newlength{\csllabelwidth}
\newlength{\cslentryspacingunit} 
\newenvironment{CSLReferences}[2] 
 {
  \setlength{\parindent}{0pt}
  \ifodd #1
  \let\oldpar\par
  \def\par{\hangindent=\cslhangindent\oldpar}
  \fi
  \setlength{\parskip}{#2\cslentryspacingunit}
 }%
 {}
\usepackage{calc}

\usepackage{xcolor}
\ifLuaTeX
  \usepackage{selnolig}  
\fi
\IfFileExists{bookmark.sty}{\usepackage{bookmark}}{\usepackage{hyperref}}
\IfFileExists{xurl.sty}{\usepackage{xurl}}{} 
\hypersetup{
  pdftitle={Language Models: A Guide for the Perplexed},
  pdfauthor={Sofia Serrano, Zander Brumbaugh, and Noah A. Smith},
  colorlinks=true,
  linkcolor={Maroon},
  filecolor={Maroon},
  citecolor={Blue},
  urlcolor={blue},
  pdfcreator={LaTeX via pandoc}}

\title{Language Models: A Guide for the Perplexed}
\author{\textbf{Sofia Serrano$^\ast$ \quad Zander Brumbaugh$^\ast$ \quad Noah A.~Smith$^{\ast\dagger}$} \\ $^\ast$Paul G.~Allen School of Computer Science \& Engineering, University of Washington \\ $^\dagger$Allen Institute for Artificial Intelligence \\ \texttt{\{sofias6,brumbzan,nasmith\}@cs.washington.edu}}
\date{}

\begin{document}
\maketitle

{
\hypersetup{linkcolor=}
\setcounter{tocdepth}{4}
\tableofcontents
}
\definecolor{lightyellow}{rgb}{1, 0.894, 0.678}
\newcommand{\takeaway}[1]{\noindent\fcolorbox{black}{lightyellow}{%
    \noindent\parbox{\linewidth - 2\fboxsep}{%
        \textbf{#1}
    }%
}}

\hypertarget{introduction}{%
\section{Introduction}\label{introduction}}

In late November 2022, OpenAI released a web-based chatbot, ChatGPT.
Within a few months, ChatGPT was reported to be the fastest-growing
application in history, gaining over 100 million users. Reports in the
popular press touted ChatGPT's ability to engage in conversation, answer
questions, play games, write code, translate and summarize text, produce
highly fluent content from a prompt, and much more. New releases and
competing products have followed, and there has been extensive
discussion about these new tools: How will they change the nature of
work? How should educators respond to the increased potential for
cheating in academic settings? How can we reduce or detect
misinformation in the output? What exactly does it take (in terms of
engineering, computation, and data) to build such a system? What
principles should inform decisions about the construction, deployment,
and use of these tools?

Scholars of artificial intelligence, including ourselves, are baffled by
this situation. Some were taken aback at how quickly these tools went
from being objects of mostly academic interest to artifacts of
mainstream popular culture. Some have been surprised at the boldness of
claims made about the technology and its potential to lead to benefits
and harms. The discussion about these new products in public forums is
often polarizing. When prompted conversationally, the fluency of these
systems' output can be startling; their interactions with people are so
realistic that some have proclaimed the arrival of human-like
intelligence in machines, adding a strong emotional note to
conversations that, not so long ago, would have mostly addressed
engineering practices or statistics.

Given the growing importance of AI literacy, we decided to write this
tutorial to help narrow the gap between the discourse among those who
study language models---the core technology underlying ChatGPT and
similar products---and those who are intrigued and want to learn more
about them. In short, we believe the perspective of researchers and
educators can add some clarity to the public's understanding of the
technologies beyond what's currently available, which tends to be either
extremely technical or promotional material generated about products by
their purveyors.

Our approach teases apart the concept of a language model from products
built on them, from the behaviors attributed to or desired from those
products, and from claims about similarity to human cognition. As a
starting point, we:

\begin{enumerate}
\def\labelenumi{\arabic{enumi}.}
\tightlist
\item
  Offer a scientific viewpoint that focuses on questions amenable to
  study through experimentation,
\item
  Situate language models as they are today in the context of the
  research that led to their development, and
\item
  Describe the boundaries of what is known about the models at this
  writing.
\end{enumerate}

Popular writing offers numerous, often thought-provoking metaphors for
LMs, including
\href{https://www.economist.com/by-invitation/2023/06/21/artificial-intelligence-is-a-familiar-looking-monster-say-henry-farrell-and-cosma-shalizi}{bureaucracies
or markets} (Henry Farrell and Cosma Shalizi),
\href{https://interhumanagreement.substack.com/p/demons}{demons} (Leon
Derczynski), and a
\href{https://www.newyorker.com/tech/annals-of-technology/chatgpt-is-a-blurry-jpeg-of-the-web}{``blurry
JPEG'' of the web} (Ted Chiang). Rather than offering a new metaphor, we
aim to empower readers to make sense of the discourse and contribute
their own. Our position is that demystifying these new technologies is a
first step toward harnessing and democratizing their benefits and
guiding policy to protect from their harms.

LMs and their capabilities are only a part of the larger research
program known as artificial intelligence (AI). (They are often grouped
together with technologies that can produce other kinds of content, such
as images, under the umbrella of ``generative AI.'') We believe they're
a strong starting point because they underlie the ChatGPT product, which
has had unprecedented reach, and also because of the immense potential
of natural language for communicating complex tasks to machines. The
emergence of LMs in popular discourse, and the way they have captured
the imagination of so many new users, reinforces our belief that the
language perspective is as good a place to start as any in understanding
where this technology is heading.

The guide proceeds in five parts. We first introduce concepts and tools
from the scientific/engineering field of natural language processing
(NLP), most importantly the notion of a ``task'' and its relationship to
data (section \ref{thinkinglikenlp}). We next define language modeling
using these concepts (section \ref{lmtask}). In short, language modeling
automates the prediction of the next word in a sequence, an idea that
has been around for decades. We then discuss the developments that led
to the current so-called ``large'' language models (LLMs), which appear
to do much more than merely predict the next word in a sequence (section
\ref{lmstollms}). We next elaborate on the current capabilities and
behaviors of LMs, linking their predictions to the data used to build
them (section \ref{capabilities}). Finally, we take a cautious look at
where these technologies might be headed in the future (section
\ref{future}). To overcome what could be a terminology barrier to
understanding admittedly challenging concepts, we also include a
Glossary of NLP and LM words/concepts (including ``perplexity,'' wryly
used in the title of this Guide).

\hypertarget{thinkinglikenlp}{%
\section{Background: Natural language processing concepts and
tools}\label{thinkinglikenlp}}

Language models as they exist today are the result of research in
various disciplines, including information theory, machine learning,
speech processing, and natural language processing.\footnote{A ``natural
  language'' is a language that developed naturally in a community, like
  Hawaiian or Portuguese or American Sign Language. For the most part,
  NLP researchers focus on human languages and specifically written
  forms of those languages. Most often, natural languages contrast with
  programming languages like Python and C++, which are artifacts
  designed deliberately with a goal in mind.} This work's authors belong
to the community of natural language processing (NLP) researchers,
members of which have been exploring the relationship between computers
and natural languages since the 1960s.\footnote{There are other uses of
  the ``NLP'' acronym with very different meanings. Ambiguous terms and
  expressions are common in natural languages, and one of the challenges
  of the field of NLP.} Two fundamental and related questions asked in
this community are: ``In what ways can computers understand and use
natural language?'' and ``To what extent can the properties of natural
languages be simulated computationally?'' The first question has been
approached mainly by attempts to build computer programs that show
language-understanding and language-use behavior (such as holding a
conversation with a person); it is largely treated as an engineering
pursuit that depends heavily on advances in hardware. The second
question brings NLP into contact with the fields of linguistics,
cognitive science, and psychology. Here, language tends to be viewed
through a scientific lens (seeking to experimentally advance the
construction of theories about natural language as an observable
phenomenon) or sometimes through a mathematical lens (seeking formal
proofs). Because these two questions are deeply interconnected, people
interested in either of them often converse and collaborate, and many
are interested in both questions.

We believe the concepts (ideas, terminology, and questions) and tools
(problem-solving methods) the NLP community uses in research are helpful
in advancing understanding of language models. They are familiar to many
AI researchers and practitioners, and similar ones have evolved in other
communities (for example, computer vision). If you have experience with
computer programming, data science, or the discrete math foundations of
computer science, you may have been exposed to these ideas before, but
we don't believe they are universally or consistently taught in classes
on those topics. Having a basic understanding of them will help you to
think like an NLP expert.

\hypertarget{taskification}{%
\subsection{Taskification: Defining what we want a system to
do}\label{taskification}}

The first step in building a machine is deciding what we want the
machine to do. People who build power plants, transportation devices, or
cooking appliances work from a \emph{specification} that spells out the
inputs and outputs of the desired system in great detail. It's not
enough to say that ``the power plant must provide electricity to all the
homes in its town.'' Engineers require a precise statement of how many
kiloWatt-hours are to be produced, the budget for building the plant,
environmental impacts expected, all the laws regulating the construction
of plants that are in effect to guarantee safety, and much more.

To take an example that's much simpler and more relevant to building an
NLP system, consider a computer program (which is a ``machine'' in a
very abstract sense; we'll also call it a ``system'') that sorts a list
of names alphabetically. This task sounds simple, and computer science
students would likely start thinking about different procedures for
sorting lists. There are, however, some details that need to be
addressed before we start writing code, such as:

\begin{itemize}
\tightlist
\item
  How will the names be input to the program, and what should the
  program do with the output? (E.g., will the program run locally on a
  user's laptop? Or is there a web interface users will use to type in
  the input and then see the output in their browser tab? Or will they
  upload/download files? If so, what is the format for those files?)
\item
  What set of characters will appear in the input, and what rules are we
  using to order them? (E.g., how do we handle the apostrophe in a name
  like ``O'Donnell''? How should diacritic (accented) characters be
  handled? What happens if some names are in Latin script and others in
  Arabic script?)
\item
  Are there constraints on how much memory the program can use, or on
  how quickly it needs to execute? If the input list is so long that the
  program will violate those constraints, should the user get a failure
  message?
\end{itemize}

\begin{figure}
\centering
\includegraphics{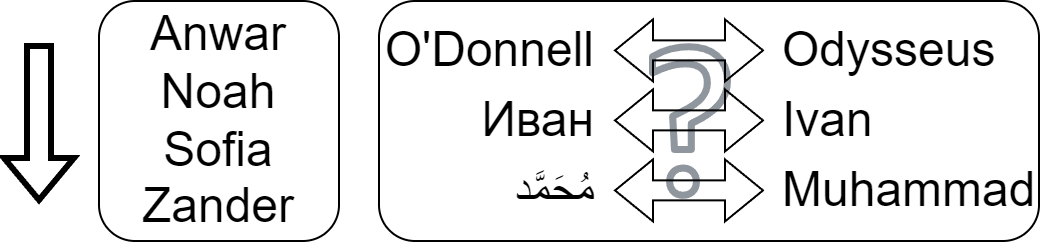}
\caption{Some tasks, like alphabetical name sorting, may seem very
simple but often raise detailed questions that must be addressed for a
full specification.}
\end{figure}

These may seem like tedious questions, but the more thoroughly we
anticipate the eventual use of the system we're building, the better we
can ensure it will behave as desired across all possible cases.

\hypertarget{abstract-vs.-concrete-system-capabilities}{%
\subsubsection{Abstract vs.~concrete system
capabilities}\label{abstract-vs.-concrete-system-capabilities}}

When building an NLP system, the situation is no different than the name
sorter, except that it's considerably harder to be precise. Consider
some of the kinds of capabilities the NLP community has been targeting
in its sixty-year history:

\begin{itemize}
\tightlist
\item
  Translate text from one language to another
\item
  Summarize one or more documents in a few paragraphs or in a structured
  table
\item
  Answer a question using information in one or more documents
\item
  Engage in a conversation with a person and follow any instructions
  they give
\end{itemize}

Each of these high-level applications immediately raises a huge number
of questions, likely many more than for simpler applications like the
name sorter, because of the open-ended nature of natural language input
(and output). Some answers to those questions could lead an expert very
quickly to the conclusion that the desired system just isn't possible
yet or would be very expensive to build with the best available methods.
Researchers make progress on these challenging problems by trying to
define \textbf{\emph{tasks}}, \emph{or versions of the application that
abstract away some details while making some simplifying assumptions.}

For example, consider the translation of text from one language to
another. Here are some fairly conventional assumptions made in many
translation research projects:

\begin{itemize}
\tightlist
\item
  The input text will be in one of a small set of languages; it will be
  formatted according to newspaper-like writing conventions. The same
  holds for the output text.
\item
  Text will be translated one sentence or relatively short segment of
  text at a time.
\item
  The whole segment will be available during translation (that is,
  translation isn't happening in ``real time'' as the input text is
  produced, as might be required when subtitling a live broadcast).
\end{itemize}

It's not hard to find research on automatic translation that makes
different assumptions from those above. A new system that works well and
relies on fewer assumptions is typically celebrated as a sign that the
research community is moving on to harder problems. For example, it's
only in the past few years that we have made the leap from systems that
support single input-to-output translations to systems that support
multiple input-to-output languages. We highlight that there are always
some narrowing assumptions, hopefully temporary, that make a problem
more precise and therefore more solvable.

We believe that many discussions about AI systems become more
understandable when we recognize the assumptions beneath a given system.
There is a constant tension between tasks that are more
general/abstract, on which progress is more impactful and exciting to
researchers, and tasks that are more specific/concrete. Solving a
concrete, well-defined task may be extremely useful to someone, but
certain details of how that task is defined might keep progress on that
task from being useful to someone else. To increase the chances that
work on a concrete task will generalize to many others, it's vital to
have a real-world user community engaged in the definition of that task.

\hypertarget{we-need-data-and-an-evaluation-method-for-research-progress-on-a-task}{%
\subsubsection{We need data and an evaluation method for research
progress on a
task}\label{we-need-data-and-an-evaluation-method-for-research-progress-on-a-task}}

The term ``task'' is generally used among researchers to refer to a
specification of certain components of an NLP system, most notably data
and evaluation:

\begin{itemize}
\tightlist
\item
  \textbf{Data}: there is a set of realistic demonstrations of possible
  inputs paired with their desirable outputs.
\item
  \textbf{Evaluation}: there is a method for measuring, in a
  quantitative and reproducible way, how well any system's output
  matches the desired output.
\end{itemize}

Considerable research activity focuses on building datasets and
evaluation methods for NLP research, and the two depend heavily on each
other. Consider again the translation example. Examples of translation
between languages are easy to find for some use cases. A classic example
is parliamentary language translated from English to French, or vice
versa. The proceedings of the Canadian Parliament are made available to
the public in both English and French, so human translators are
constantly at work producing such demonstrations; paired bilingual texts
are often called ``parallel text'' in the research community. The
European Parliament does the same for multiple languages. Finding such
data isn't as easy for some languages or pairs of languages, and as a
result, there has been considerably more progress on automated
translation for European languages than for others.

What about evaluation of translation? One way to evaluate how well a
system translates text is to take a demonstration, feed the input part
to a system, and then show a human judge the desired output and the
system output. We can ask the judge how faithful the system output is to
the desired output. If the judge speaks both languages, we can show them
the input instead of the desired output (or in addition to it) and ask
the same question. We can also ask human judges to look only at the
system output and judge the fluency of the text. As you can imagine,
there are many possible variations, and the outcomes might depend on
exactly what questions we ask, how we word those questions, which judges
we recruit, how much they know about translation systems already, how
well they know the language(s), and whether and how much we pay them.

In 2002, to speed up translation evaluation in research work,
researchers introduced a fully automated way to evaluate translation
quality called ``Bleu'' scores
(\protect\hyperlink{ref-papineni-etal-2002-bleu}{Papineni et al. 2002}),
and there have been many proposed alternatives since then, with much
discussion over how well these cheaper automatic methods correlate with
human judgments. One challenge for automatic evaluation of translation
is that natural languages offer many ways to say the same thing. In
general, reliably rating the quality of a translation could require
recognizing all of the alternatives because the system could (in
principle) choose any of them.

We used translation as a running example precisely because these
questions are so contentious and potentially costly for this task. We'll
next consider a fairly concrete task that's much simpler: categorizing
the overall tone of a movie review (positive vs.~negative),
instantiating a more general problem known as \textbf{sentiment
analysis}. Here, researchers have collected demonstrations from movie
review websites that pair reviews with numerical ratings (e.g.,, one to
five stars). If a system takes a review as input and predicts the
rating, we can easily check whether the output exactly matches the
actual rating given by the author, or we could calculate the difference
between the system and correct ratings. Here, the collection of data is
relatively easy, and the definition of system quality is fairly
uncontroversial: the fewer errors a system makes (or the smaller the
difference between the number of author stars and system-predicted
stars), the higher the system's quality.

Note, however, that a system that does well on the movie review
sentiment task may not do so well on reviews of restaurants, electronics
products, or novels. This is because the language people use to say what
they like or don't like about a movie won't carry the same meaning in a
different context. (If a reviewer says that a movie ``runs for a long
time,'' that isn't as obviously positive as the same remark about a
battery-operated toothbrush, for example.) In general, knowing the scope
of the task and how a system was evaluated are crucial to understanding
what we can expect of a system in terms of its generalizability, or how
well its performance quality holds up as it's used on inputs less and
less like those it was originally evaluated on. It's also essential when
we compare systems; if the evaluations use different demonstrations or
measure quality differently, a comparison won't make sense.

For most of its history, NLP has focused on research rather than
development of deployable systems. Recent interest in user-facing
systems highlights a longstanding tension in taskification and the
dataset and evaluation requirements. On one hand, researchers prefer to
study more abstract tasks so that their findings will be more generally
applicable across many potential systems. The scientific community will
be more excited, for example, about improvements we can expect will hold
across translation systems for many language pairs (rather than one) or
across sentiment analysis systems for many kinds of reviews (rather than
just movies). On the other hand, there is near-term value in making a
system that people want to use because it solves a specific problem
well, which requires being more concrete about the intended users, their
data, and meaningful evaluation.

There is yet another step between researching even fairly concrete tasks
and building usable systems. These are evaluated very differently.
Evaluations in research tend to focus on specific, narrowly defined
capabilities, as exemplified in a sample of data. It's an often unstated
assumption in research papers that improved task performance will
generalize to similar tasks, perhaps with some degradation. The research
community tends to share such assumptions, with the exception of
research specifically on generalization and robustness across domains of
data. Meanwhile, deployable systems tend to receive more rigorous
testing with intended users, at least to the extent that they are built
by organizations with an interest in pleasing those users. In
deployment, ``task performance'' is only part of what's expected
(systems must also be reasonably fast, have intuitive user interfaces,
pose little risk to users, and more).

\noindent\fcolorbox{black}{lightyellow}{%
    \noindent\parbox{\linewidth - 2\fboxsep}{%
        \textbf{People interested in NLP systems should be mindful of the gaps between (1) high-level, aspirational capabilities, (2) their "taskified" versions that permit measurable research progress, and (3) user-facing products. As research advances, and due to the tension discussed above, the "tasks" and their datasets and evaluation measures are always in flux.}
    }%
}

\hypertarget{closerlookatdata}{%
\subsection{A closer look at data: where it comes from and how it's
used}\label{closerlookatdata}}

For the two task examples discussed above (translation and sentiment
analysis tasks), we noted that demonstrations (inputs with outputs)
would be relatively easy to find for some instances of the tasks.
However, data might not always be so easy to come by. The availability
of data is a significant issue for two reasons:

\begin{itemize}
\tightlist
\item
  For most NLP applications, and most tasks that aim to approximate
  those applications, there is no ``easy'' source of data. (Sentiment
  analysis for movie reviews is so widely studied, we believe, because
  the data is unusually easy to find, not because there is especially
  high demand for automatic number-of-stars prediction.)
\item
  The best known techniques for building systems require access to
  substantial amounts of extra data to \emph{build} the system, not just
  to evaluate the quality of its output.
\end{itemize}

\hypertarget{differentiating-training-from-test-data}{%
\subsubsection{Differentiating training from test
data}\label{differentiating-training-from-test-data}}

From here on, we refer to data used to build a system as
\textbf{training data} and data used to evaluate systems as \textbf{test
data}. This distinction is extremely important for a reason that's easy
to understand.

\begin{figure}
\centering
\includegraphics{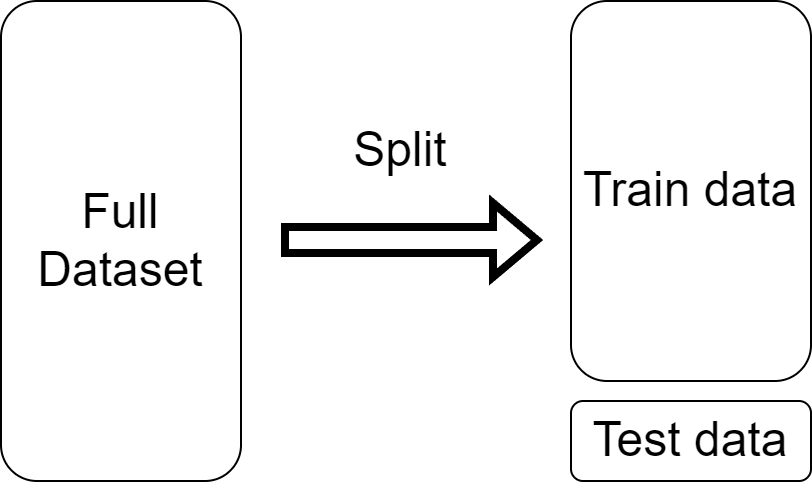}
\caption{When data is split into training and test sets, it's critical
there is no overlap between the two.}
\end{figure}

Consider a student who somehow gets a copy of the final exam for one of
their classes a few weeks before the exam. Regardless of how much the
student is to blame in accessing the test, regardless of whether they
even knew the exam they saw was the actual final exam, regardless of how
honorably they behaved during those weeks and during the test, if they
get a high score, the instructor cannot conclude that the student
learned the material. The same holds true for an NLP system. For the
test data to be useful as an indicator of the quality of the system's
output, it is necessary that the test data be ``new'' to the system. We
consider this the cardinal rule of experimentation in NLP:
\textbf{\emph{The test data cannot be used for any purpose prior to the
final test.}} Occasionally, someone will discover a case where this rule
was violated, and (regardless of the intent or awareness of those who
broke the rule) the conclusions of any research dependent on that case
must be treated as unreliable.

\noindent\fcolorbox{black}{lightyellow}{%
    \noindent\parbox{\linewidth - 2\fboxsep}{%
        \textbf{To get a sense of an NLP system's actual quality, it is crucial that the system not be evaluated on data it has seen during training.}
    }%
}

\hypertarget{creating-a-dataset-from-scratch}{%
\subsubsection{Creating a dataset from
scratch}\label{creating-a-dataset-from-scratch}}

Let's consider a variant of the sentiment analysis problem that might
emerge in a high-stakes academic decision-making setting. Suppose we
plan to build an NLP system that reads recommendation letters for
applicants to a university degree program. The system should rate the
sentiment of the recommender toward the applicant. On the surface, this
is similar to the movie review problem we discussed previously. But this
use case introduces some new challenges.

First, we are unlikely to find demonstrations that we could use to train
or evaluate a system.\footnote{In NLP terms, \emph{finding} and
  collecting such existing demonstrations would count as dataset
  creation. ``Creating a dataset'' in NLP can refer to either creating
  of new text via expert annotation or crowdsourcing, \emph{or}
  collecting existing text into a more readily accessible form for model
  developers, such as via web crawling or scraping.} Recommendation
letters are extremely private; those who write them do so on the
assumption that they will not be revealed to anyone who doesn't need to
read them to assess the application. If we manage to find recommendation
letters on the public web, it's likely that they either aren't supposed
to be there (and are therefore unethical to use) or they're synthetic
examples used to teach people how to write or evaluate recommendation
letters (and therefore artificial and probably different from actual
letters in key practical ways---remember that we need \emph{realistic}
demonstrations).

A second issue is that the information conveyed in a recommendation
letter is often complex, considering many aspects of a candidate's
performance and potential. Mapping the letter down to a single number or
category seems quite challenging (if it were easy, we wouldn't ask
recommenders to write letters, we'd only ask them to report the number
or category). Finally, as anyone who has been on an admissions or hiring
committee knows, there is a great deal of subjectivity in
\emph{interpreting} a recommendation letter. Different readers may draw
different conclusions about the prevailing signal in a single letter.
Even if we overcome the hurdle of finding letters to use, that's only
half of what we need because the demonstrations need to include desired
\emph{outputs} as well as inputs.

Indeed, the tasks that researchers explore or system builders try to
explore are very often limited by the data that's available. When the
desired data (or anything similar to it) is unavailable, it's sometimes
possible to create it. For example, to automate sentiment analysis of
social media messages about a particular much-discussed public figure,
we could hire people to do the task of labeling a sample of messages,
essentially demonstrating the desired behavior for our eventual system.
Labeling tweets about a politician might be relatively easy for someone
who speaks the language of the tweets and is familiar with the social
context.

Some tasks, in contrast, require much more expertise. For example, to
build a system that answers questions about medical journal articles,
we'd want the data to be created by people who know how to read and
understand such articles so that the answers are accurate and grounded
in article specifics. Of course, experts will be more costly to employ
for this work than non-experts. A major tradeoff in the creation of
datasets for NLP is between the inherent quality and diversity of the
demonstrations and the cost of producing them. We believe that
high-quality data is \emph{always} essential for reliable evaluations
(test data) and \emph{usually} essential for high performance on those
evaluations (training data).

\noindent\fcolorbox{black}{lightyellow}{%
    \noindent\parbox{\linewidth - 2\fboxsep}{%
        \textbf{Collecting training data for most NLP tasks is quite difficult, and this often impacts which possible NLP applications or problems are studied.}
    }%
}

\hypertarget{task-data}{%
\subsection{Building an NLP system}\label{task-data}}

For almost a decade, and with a small number of exceptions, the dominant
approach to building an NLP system for a particular task has been based
on machine learning. \textbf{Machine learning (ML)} refers to a body of
theoretical and practical knowledge about data-driven methods for
solving problems that are prohibitively costly for humans to solve.
These methods change over time as new discoveries are made, as different
performance requirements are emphasized, and as new hardware becomes
available. A huge amount of tutorial content is already available about
machine learning methods, with new contributions following fast on the
heels of every new research advance. Here, we introduce a few key ideas
needed to navigate the current landscape.

The first concept is a \textbf{parameter.} A parameter is like a single
knob attached to a system: Turning the knob affects the behavior of the
system, including how well it performs on the desired task. To make this
concrete, let's consider an extremely simple system for filtering spam
emails. Due to budgetary constraints, this system will have only one
parameter. The system works as follows: it scans an incoming email and
increments a counter every time it encounters an ``off-color'' word
(e.g., an instance of one of the seven words the comedian George Carlin
claimed he wasn't allowed to say on television). If the count is too
high, the email is sent to the spam filter; otherwise, it goes to the
inbox. How high is too high? We need a threshold, and we need to set it
appropriately. Too high, and nothing will get filtered; too low, and too
many messages may go to spam. The threshold is an example of a
parameter.

This example neatly divides system-building problem into two separate
parts:

\begin{enumerate}
\def\labelenumi{\arabic{enumi}.}
\tightlist
\item
  \textbf{Deciding what parameters the system will have and how they
  will work.} In our spam example, the system and the role of the
  off-color word threshold parameter are easy to explain. The term
  \textbf{architecture} (or model architecture, to avoid confusion with
  hardware architecture) typically refers to the decision about what
  parameters a model will have. For example, picture a generic-looking
  black box with lots of knobs on it; the box has a slot on one side for
  inputs and a slot on the other side for outputs. The ``architecture''
  of that model refers to the number of knobs, how they're arranged on
  the box, and how their settings affect what occurs inside the box when
  it turns an input into an output.
\item
  \textbf{Setting parameter values.} This corresponds to determining
  what value each individual knob on the box is turned to. While we
  likely have an intuition about how to set the parameter in the spam
  example, the value that works the best is probably best determined via
  experimentation.
\end{enumerate}

We now walk through how ML works in more detail and introduce some
components you'll likely hear about if you follow NLP developments.

\hypertarget{architectures-neural-networks}{%
\subsubsection{Architectures: Neural
networks}\label{architectures-neural-networks}}

Today, the vast majority of architectures are \textbf{neural networks}
(sometimes called \emph{artificial} neural networks to differentiate
them from biological ones). For our purposes, it's not important to
understand what makes neural networks special as a category of
architectures. However, we should know that their main properties
include (1) large numbers of parameters (at this writing, trillions) and
(2) being differentiable\footnote{We are referring to the concept from
  calculus. If a function is ``differentiable'' with respect to some
  numbers it uses, then calculus gives us the ability to calculate which
  \emph{small} changes to those variables would result in the biggest
  change to the function.} functions with respect to those parameters:
addition, subtraction, exponentiation, trigonometric functions, etc.,
and combinations of them. A general observation about neural network
architectures (but not a necessary or defining property) is that the
relationship between their numerical calculations and the task-solving
behavior of a model (after its parameters are set) is not explainable to
human observers. This is why they are associated with the metaphor of a
\emph{black box} (whose internal components can't be observed or easily
understood).

\hypertarget{minimization}{%
\subsubsection{Choosing values for all the parameters: Minimizing a loss
function}\label{minimization}}

In order to work well, a neural network needs to have its parameters set
to useful values (i.e., values that will work well together to
mathematically transform each input into an output close to the input's
correct answer). But how do we choose parameters' values when we have so
many we need to decide? In this section, we describe the general
strategy that we use in NLP.

Imagine yourself in the following (admittedly not recommended) scenario.
At night, and with no GPS or source of light on you, you are dropped in
a random location somewhere over the Cascade Range in Washington State
with the instructions to find the deepest valley you can (without just
waiting for morning). You move your feet to estimate the steepest
downward direction. You take a small, careful step in that direction and
repeat until you seem to be in a flat place where there's no direction
that seems to take you farther downward.

Machine learning (and, by extension, NLP) views the setting of parameter
values as a problem of \textbf{numerical optimization}, which has been
widely studied for many years by mathematicians, statisticians,
engineers, and computer scientists. One of the tools of machine learning
is an automated procedure that frames the parameter value-setting
problem like that terrifying hike. Recall that we said that neural
networks need to be \emph{differentiable} with respect to their
parameters--- that is, they need to be set up to allow calculus to tell
us which tiny change to each parameter will result in the steepest
change of \emph{something} calculated using the neural network's output.
In our nighttime hike scenario, at each step, we make a tiny adjustment
to our north-south and east-west coordinates (i.e., position on the
map). To adjust the parameters of our neural network, we will consider
our current set of parameters our ``coordinates'' and likewise
repeatedly make tiny adjustments to our current coordinates. But what
does it mean to move ``down'' in this context? Ideally, moving ``down''
should correspond to our neural network producing outputs that better
match our data. How can we define a function---our ``landscape''--- such
that this is true?

A \textbf{loss function} is designed for precisely this purpose: to be
lower when a neural network performs better. In short, a loss function
evaluates how well a model's output resembles a set of target values
(our training data), with a higher ``loss'' signifying a higher error
between the two. The more dissimilar the correct output is from the
model's produced output, the higher the loss value should be; if they
match, it should return zero. This means a loss function should ideally
be closely aligned to our evaluation method.\footnote{You can think of a
  loss function as a stern, reserved teacher grading a student's work.
  The student (the model whose parameters we want to set) is given an
  exam question (an input to the model) and produces an answer. The
  teacher mechanically compares the question's correct answer to the
  student's answer, and then reports how many points have been deducted
  for mistakes. When the student gets the answer perfectly right, the
  loss will be zero; no points are deduced. We discuss some additional
  mathematical details of loss functions in the appendix.}

By performing the following procedure, we are able to train a
neural-network-based model:

\begin{enumerate}
\def\labelenumi{\arabic{enumi}.}
\tightlist
\item
  We use a loss function to define our landscape for our model's
  nighttime hike based on our training inputs and outputs,
\item
  we make a small adjustment to each of our coordinates (model
  parameters) to move ``down'' that landscape towards closer matches
  between our model's outputs and the correct ones, and
\item
  we repeat step 2 until we can't make our model's outputs any more
  similar to the correct ones.
\end{enumerate}

This method is known as \textbf{(stochastic) gradient descent (SGD),}
since the direction that calculus gives us for each parameter is known
as the ``gradient.''

Leaving aside some important details (for example, how to efficiently
calculate the gradients using calculus, working out precisely when to
stop, exactly how much to change the parameter values in step 3, and
some tricks that make the algorithm more stable), this method has proven
effective for choosing parameter values in modern model architectures
and in their predecessors.

\hypertarget{the-hardware-graphics-processing-units-gpus}{%
\subsubsection{The hardware: Graphics processing units
(GPUs)}\label{the-hardware-graphics-processing-units-gpus}}

For over a decade, graphics processing units (GPUs) have been the main
type of hardware used to train NLP models based on neural networks. This
may seem counterintuitive (since it's language we're processing here,
not graphics). However, GPUs are effective for doing many matrix and
vector calculations in parallel, and successful neural network
architectures have used these parallel calculations to perform
input-to-output mapping quickly (since stochastic gradient descent
requires that mapping to be performed many many times during training).
Indeed, the realization that neural networks were well-suited to train
on GPUs proved to be crucial to their widespread adoption.

\hypertarget{lmtask}{%
\section{The language modeling task}\label{lmtask}}

Section \ref{thinkinglikenlp} introduced some NLP concepts and tools,
including the idea of encapsulating a desired application into a
``task,'' the importance of datasets, and a high-level tour of how
systems learn to perform a task using data. Here, we turn to language
modeling, a specific task.

\hypertarget{lmnextword}{%
\subsection{Language modeling as next word
prediction}\label{lmnextword}}

The language modeling task is remarkably simple in its definition, in
the data it requires, and in its evaluation. Essentially, its goal is to
predict the next word in a sequence (the output) given the sequence of
preceding words (the input, often called the ``context'' or ``preceding
context''). For example, if we ask you to come up with an idea of which
word might come next in a sentence in progress---say, ``This document is
about Natural Language \_\_\_\_''---you're mentally performing the
language modeling task. The real-world application that should come to
mind is some variation on an auto-complete function, which at this
writing is available in many text messaging, email, and word processing
applications.

Language modeling was for several decades a core component in systems
for speech recognition and text translation. Recently, it has been
deployed for broad-purpose conversational chat, as in the various GPT
products from OpenAI, where a sequence of ``next'' words is predicted as
a sequential response to a natural language prompt from a user.

\begin{figure}
\centering
\includegraphics{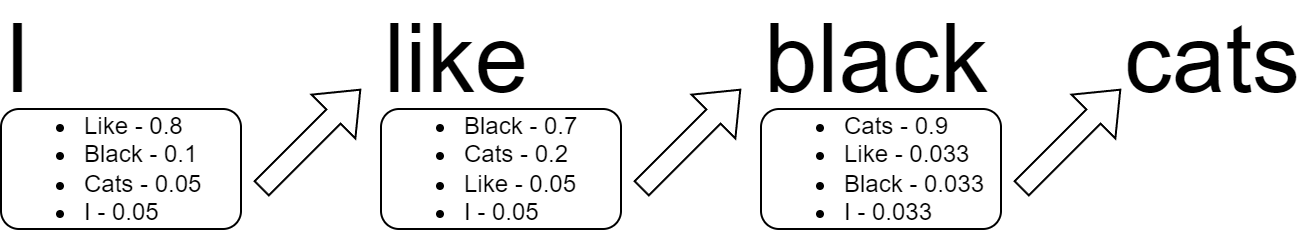}
\caption{Next word prediction samples a word from the language model's
guess of what comes next at each time step.}
\end{figure}

What would make it possible to achieve high accuracy at predicting the
next word across many contexts? At a fundamental level, natural language
is predictable because it is highly structured. People unconsciously
follow many rules when they use language (e.g., English speakers mostly
utter verbs that agree with their subjects sometime \emph{after} those
subjects, and they place adjectives \emph{before} the nouns whose
meaning they modify). Also, much of our communication is about
predictable, everyday things (consider how frequently you engage in
small talk).

As an NLP task, language modeling easily checks the two critical boxes
we discussed in section \ref{thinkinglikenlp}: data and evaluation. LMs
need only text; every word in a large collection of text naturally comes
with the preceding context of words. When we say ``only text,'' we mean
specifically that we don't need any kind of label to go with pieces of
text (like the star ratings used in sentiment analysis tasks, or the
human-written translations used in translation tasks). The text itself
is comprised of inputs \emph{and} outputs. Because people produce text
and share it in publicly visible forums all the time, the amount of text
available (at least in principle, ignoring matters of usage rights) is
extremely large. The problem of fresh, previously unseen test data is
also neatly solved because new text is created every day, reflecting new
events and conversations in the world that are reliably different from
those that came before. There is also a relatively non-controversial
evaluation of LMs that requires no human expertise or labor, a more
technical topic that we return to in section \ref{perplexity}.

\hypertarget{whylm}{%
\subsection{Why do we care about language modeling?}\label{whylm}}

We have thus far established what the language modeling task is.
However, we haven't explained why this task is worth working on. Why do
we bother building a model that can predict the next word given the
words that have come before? If you already make use of auto-complete
systems, you have an initial answer to this question. But there are more
reasons.

For many years, NLP researchers and practitioners believed that a good
language model was useful only for estimating fluency. To illustrate
this, imagine a language model faced with guessing possible
continuations for a partial sentence like ``The dog ate the \_\_\_\_''
or ``Later that afternoon, I went to a \_\_\_\_.'' As English speakers,
we share a pretty strong sense that the following word is likely to be
either a noun or part of a descriptor preceding a noun. Likewise, if we
have a good language model for this type of English, that model will
have needed to implicitly learn those kinds of fluency-related rules to
perform the language modeling task well. This is why LMs have
historically been incorporated as a component in larger NLP systems,
such as machine translation systems; by taking their predictions (at
least partially) into account, the larger system is more likely to
produce more fluent output.

In more recent years, our understanding of the value of LMs has evolved
substantially. In addition to promoting fluency, a sufficiently powerful
language model can implicitly learn a variety of world knowledge.
Consider continuations to the following partial sentences: ``The
Declaration of Independence was signed by the Second Continental
Congress in the year \_\_\_\_,'' or ``When the boy received a birthday
gift from his friends, he felt \_\_\_\_.'' While there are any number of
fluent continuations to those sentences---say, ``1501'' or ``that the
American Civil War broke out'' for the first, or ``angry'' or ``like
going to sleep'' for the second---you likely thought of ``1776'' as the
continuation for the first sentence and a word like ``happy'' or
``excited'' for the second. Why? It is likely because you were engaging
your knowledge of facts about history as well as your common sense about
how human beings react in certain situations. This implies that to
produce those continuations, an LM would need at least a rudimentary
version of this information.

\noindent\fcolorbox{black}{lightyellow}{%
    \noindent\parbox{\linewidth - 2\fboxsep}{%
        \textbf{To do a good job of guessing the continuations of text, past a certain point, an LM must have absorbed some additional kinds of information to progress beyond simple forms of fluency.}
    }%
}

NLP researchers got an early glimpse of this argument in Peters et al.
(\protect\hyperlink{ref-elmo}{2018}). This paper reported that systems
that trained an LM first as an early stage of building systems for
varied tasks, ranging from determining the answer to a question based on
a given paragraph to determining which earlier entity a particular
pronoun was referencing, far outperformed their analogous versions that
weren't informed by an LM (as measured by task-specific definitions of
quality). This finding led to widespread researcher acceptance of the
power of ``pretraining'' a model to perform language modeling and then
``finetuning'' it (using its pretrained parameters as a starting point)
to perform a non-language-modeling task of interest, which also
generally improved end-task performance.

It shouldn't be too surprising that LMs can perform well at filling in
the blanks or answering questions when the correct answers are in the
training data. For a new task, it seems that the more similar its inputs
and outputs are to examples in the pretraining data, the better the LM
will perform on that task.

\hypertarget{datanuances}{%
\subsection{Data for language models: Some nuances}\label{datanuances}}

There are two important caveats to our earlier claim that collecting
data for a language model is ``easy.'' First, because there is a massive
amount of text available on the internet which could be downloaded and
used to build or evaluate LMs, at least for research purposes, \emph{a
language model builder must decide which data to include or exclude.}
Typical sources of data include news articles, books, Wikipedia, and
other web text that is likely to be carefully edited to conform to
professional writing conventions. Some LMs include more casual text from
social media websites or online forums, or more specialized language
from scientific texts. While researchers have generally considered
training language models on publicly available text data to be covered
by fair use doctrine, the relationship between copyright protections and
language model practices is not fully settled; we discuss this further
in section \ref{emergingregulation}.

A major decision is whether to filter texts to only certain
languages.\footnote{The problem of assigning a language identifier to a
  text (e.g., is it English, Spanish, etc.?) constitutes another family
  of NLP tasks. It's a useful exercise to consider how to select the set
  of language names to use as labels for language identification,
  e.g.,which dialects of a language are separate from each other and
  should receive different labels?} Depending on the community of users
one intends the LM to serve, it may be preferable to filter text on
certain topics (e.g., erotica) or text likely to contain offensive
content or misinformation. Today's LM datasets are too large for a
person to read in a single lifetime, so automated tools are employed to
curate data. The implications of these decisions are a major topic for
current research, and we return to them in section
\ref{trainingdatacontents}.

The other caveat is a more technical one: \emph{what counts as a
``word''?} For languages with writing systems that use whitespace to
separate words, like English, this is not a very interesting question.
For writing systems with less whitespace between words (e.g., Chinese
characters), segmentation into words could be a matter of choosing an
arbitrary convention to follow or of adopting one of many competing
linguistic theories. Today, LMs are often built on text from more than
one natural language as well as programming language code. The dominant
approach to defining where every word in the data starts and ends is to
apply a fully automated solution to create a vocabulary (set of words
the language model will recognize as such) that is extremely robust
(i.e., it will \emph{always} be able to break a text into words
according to its definition of words). The approach
(\protect\hyperlink{ref-sennrich-etal-2016-neural}{Sennrich, Haddow, and
Birch 2016}) can be summed up quite simply:

\begin{itemize}
\tightlist
\item
  Any single character is a word in the vocabulary. This means that the
  LM can handle any entirely new sequence of characters by default by
  treating it as a sequence of single-character words.
\item
  The most frequently occurring two-word sequence is added to the
  vocabulary as a new word. This rule is applied repeatedly until a
  target vocabulary size is reached.
\end{itemize}

This data-driven approach to building a language modeling vocabulary is
effective and ensures that common words in the LM's training data are
added to its vocabulary. Other, rarer words will be represented as a
sequence of word pieces in the model's vocabulary (similarly to how you
might sound out an unfamiliar word and break it down into pieces you've
seen before). However, note that a lot depends on the data through the
calculation of what two-word sequence is most frequent in that data at
each step. Unsurprisingly, if the dataset used to build the vocabulary
includes little or no text from a language (or a sub-language), words in
that language will get ``chopped up'' into longer sequences of short
vocabulary words (some a single character), which has been shown to
affect how well the LM performs on text in that language.

\hypertarget{perplexity}{%
\subsection{Evaluating LMs: Perplexity}\label{perplexity}}

We mentioned earlier that the language modeling task has a
straightforward evaluation method. At first, we might think that a
``good'' language model has a low word error rate: when it guesses the
next word in a sequence, it should seldom predict the wrong word. (A
``wrong word'' here means anything other than the actual next word in
the test data sequence.)\footnote{We give a formal mathematical
  definition of word error rate in the appendix.}

LMs have generally not used the error rate to evaluate LM quality for
two reasons. First, applications sometimes predict a \emph{few} options
for the next word; perhaps it's just as good to rank the correct next
word second or third as it is to rank it first. The error rate could be
modified to count as mistakes only the cases where the correct word is
ranked below that cutoff. But how long the list should be is a question
for application designers and moves the task definition in a more
specialized/concrete direction, perhaps unnecessarily. Second, at least
earlier in the history of language modeling, most systems weren't good
enough at predicting the next word to have error rates that weren't
extremely high. If all LMs achieve error rates close to one, the error
rate measurement isn't very helpful for comparing them.

The evaluation method that \emph{is} typically used for LMs avoids both
of these issues. This method is known as \textbf{perplexity,} and can be
considered a measure of an LM's ``surprise'' as expressed through its
outputs in next word prediction. Perplexity manages to work around the
problems we've described by taking advantage of how LMs decide on a next
word in practice.

When an LM produces a next word, that next word is in reality a somewhat
processed version of that LM's actual output. What the LM
\emph{actually} produces given some input text is a \emph{probability
distribution} over its vocabulary for which word comes next. In other
words, for every possible next word in its vocabulary, the LM generates
a number between 0 and 1 representing its estimate of how likely that
word is as the continuation for the input text.\footnote{Because this is
  a probability distribution, all those numbers must add up to 1, and in
  practice, LMs always set their probabilities to numbers strictly
  greater than 0.}

Rather than evaluating an LM based on however an application developer
chooses to process those probability distributions into next words
(whether by sampling, or by choosing the word with the highest estimated
likelihood, or something else), perplexity instead directly evaluates
the probability distributions produced by the LM. Given a test set of
text, perplexity examines \emph{how high the LM's probabilities are for
the} \textbf{\emph{true}} \emph{observed next words overall,} averaged
over each word in the text-in-progress. The higher that LM's average
probability for the true words is, the \emph{lower} the LM's perplexity
(corresponding to the LM being less ``surprised'' by the actual
continuations of the text).\footnote{For those interested, we walk
  through the mathematics underlying the definition of perplexity in the
  appendix.}

Like any evaluation method, perplexity depends heavily on the test data.
In general, the more similar the training and test data, the lower we
should expect the text data perplexity to be. And if we accidentally
break the cardinal rule and test on data that was included in the
training data, we should expect extremely low perplexity (possibly
approaching 1, which is the lowest possible value of perplexity, if the
model were powerful enough to memorize long sequences it has seen in
training).

Finally, it's worth considering when perplexity seems ``too'' low. The
idea that there is some limit to this predictability, that there is
always some uncertainty about what the next word will be, is an old one
(\protect\hyperlink{ref-shannonsgame}{Shannon 1951}), motivating much
reflection on (1) how much uncertainty there actually is, and (2) what
very low perplexity on language modeling implies. Some have even
suggested that strong language modeling performance is indicative of
artificially intelligent behavior. (We return to this question in
section \ref{capabilities}.)

\hypertarget{buildinglms}{%
\subsection{Building language models}\label{buildinglms}}

Given the tools from section \ref{thinkinglikenlp} and our presentation
of the language modeling task, it's straightforward to describe how
today's best LMs are built:

\begin{enumerate}
\def\labelenumi{\arabic{enumi}.}
\tightlist
\item
  Acquire a substantial amount of diverse training data (text),
  filtering to what you believe will be high quality for your eventual
  application. Set aside some data as the test data.
\item
  Build a vocabulary from the training data.
\item
  Train a model with learnable parameters to minimize perplexity on the
  training data using a variant of stochastic gradient descent.
\item
  Evaluate the perplexity of the resulting language model on the test
  set. In general, it should be very possible to evaluate the LM on
  another test set because (1) we can check that the new proposed test
  data doesn't overlap with the training data, and (2) the vocabulary is
  designed to allow any new text to be broken into words.
\end{enumerate}

The third step reveals another attractive property of perplexity: it can
serve as a loss function because it is differentiable with respect to
the model's parameters.\footnote{In practice, the loss function is
  usually the logarithm of perplexity, a quantity known as
  cross-entropy.} Note the difference between training set perplexity
(calculated using training data) and test set perplexity calculated in
the last step.\footnote{One common question about language models is why
  they sometimes ``hallucinate'' information that isn't true. The fact
  that next word prediction is the training objective used for these
  models helps to explain this. The closest an LM comes to encoding a
  ``fact'' is through its parameters' encoding of which kinds of words
  tend to follow from a partially written sequence. Sometimes, the
  context an LM is prompted with is sufficient to surface facts from its
  training data. (Imagine our example from earlier: ``The Declaration of
  Independence was signed by the Second Continental Congress in the year
  \_\_\_\_.'' If an LM fills in the year ``1776'' after being given the
  rest of the sentence as context, that fact has been successfully
  surfaced.) Other times, however, it's not, and we just get a
  fluent-sounding next word prediction that's not actually true, or a
  ``hallucination.''}

The preceding process is how some well-known models, like GPT-2, GPT-3,
and LLaMA, were built, and it's the first step to building more recent
models like ChatGPT and GPT-4. These newer models have been further
trained on additional kinds of data (which is less ``easy'' to obtain
than the text we use for next word prediction). We return to this topic
in section \ref{adaptingasproducts}.

\hypertarget{lmstollms}{%
\section{From LMs to large language models (LLMs)}\label{lmstollms}}

Everything we've described thus far has been established for over a
decade, and some concepts much longer. Why have language models become a
topic of mainstream public conversation only recently?

Recall that a longstanding use of LMs was to estimate the fluency of a
piece of text (\ref{whylm}), especially to help text-generating systems
produce more fluent output. Only since around 2020 have LMs been
producing highly fluent output \emph{on their own,} that is, without
incorporating some other components. At this writing, you could observe
something like the text generation performance of older LMs by looking
at the autocomplete functions in messaging applications on smartphones.
If you have one of these on hand, try starting a sentence and then
finishing the sentence by picking one of the most likely next words the
autocomplete program suggests. You're likely to notice that while the
short-term continuations to the sentence are reasonable, the text
quickly devolves into moderately fluent incoherence, nothing like text
produced by state-of-the-art web-based products.

Having established the foundations---the language modeling task and the
basic strategy for building a language model---we'll now consider the
factors that have recently transformed the mostly academic language
models of the last decade into the so-called large language models
(LLMs) of today.

\hypertarget{the-move-towards-more-data}{%
\subsection{The move towards more
data}\label{the-move-towards-more-data}}

This is not a history book, but there is one obvious lesson to be
learned from the history of NLP: more training data helps make higher
quality models. One period of major changes in the field occurred in the
late 1980s and 1990s when three trends converged almost concurrently:

\begin{enumerate}
\def\labelenumi{\arabic{enumi}.}
\tightlist
\item
  Increasingly large collections of naturalistic, digital text data
  became easier to access by growing numbers of researchers thanks to
  the rise of the internet and the world-wide web.
\item
  Researchers shifted from defining rules for solving NLP tasks to using
  statistical methods that depend on data. This trend came about in part
  due to interaction with the speech processing community, which began
  using data-driven methods even earlier.
\item
  Tasks, as we described them above, became more mature and
  standardized, allowing more rigorous experimental comparisons among
  methods for building systems. This trend was driven in part by
  government investment in advancing NLP technology, which in turn
  created pressure for quantitative measures of progress.
\end{enumerate}

During the 1990s and 2000s, the speed of progress was higher for tasks
where the amount of available training data increased the fastest.
Examples include topic classification and translation among English,
French, German, and a few other languages. New tasks emerged for which
data was easy to get, like sentiment analysis for movies and products
sold and reviewed online. Meanwhile, progress on tasks where data was
more difficult to obtain (such as long text summarization, natural
language interfaces to structured databases, or translation for language
pairs with less available data) was slower. In particular, progress on
NLP for English tasks was faster than for other languages, especially
those with relatively little available data.

The recognition that more data tends to help make better systems
generates a lot of enthusiasm, but we feel obliged to offer three
cautionary notes. First, easily available data for a task doesn't make
that task inherently worth working on. For example, it's very easy to
collect news stories in English. Because the style of many
English-language newspapers puts the most important information in the
first paragraph, it's very easy to extract a decent short summary for
each story, and we now have a substantial number of demonstrations for
an English-language news summarizer. However, if readers of the news
already know that the first paragraph of a news story is usually a
summary, why build such a system? We should certainly not expect a
system built on news summarization task data to carry over well to tasks
that require summarizing scientific papers, books, or laws.

The second cautionary note is that the \emph{lack} of easy data for a
task doesn't mean the task \emph{isn't} worth solving. Consider a
relatively isolated community of people who have more recently gained
access to the internet. If they do not speak any of the dominant
languages on the internet, they may be unable to make much use of that
access. The relative absence of this community's language from the web
is one reason that NLP technology will lag behind for them. This
inequity is one of the drawbacks of data-driven NLP.

The third cautionary note is that data isn't the \emph{only} factor in
advancing NLP capabilities. We already mentioned evaluation methods. But
there are also algorithms and hardware, both of which have changed
radically over the history of NLP. We won't go into great detail on
these technical components here, but we note that the suitability of an
algorithm or a hardware choice for an NLP task depends heavily on the
quality and quantity of training data. People often use the term
``scale'' to talk about the challenges and opportunities associated with
very large training datasets. As early as 1993, researchers were
claiming that ``more data is always better data''
(\protect\hyperlink{ref-church-mercer-1993-introduction}{Church and
Mercer 1993}). We would add that which algorithms or computers are
better for building a system that performs a task depends highly on the
availability of appropriate data for that task, whether high or low or
in between. And indeed, as it turns out, the second factor we now
mention falls into the category of a change in algorithm: a change in
model architecture.

\hypertarget{transformers}{%
\subsection{The architecture: Transformers}\label{transformers}}

Not long ago, students of NLP would be introduced to a wide range of
different architectures. One would likely hear about the relative merits
of each and learn what particular kinds of problems it was well suited
to solve. From year to year, new ones would be added, sometimes
replacing those no longer deemed optimal in any setting. Today, these
diverse architectures have virtually all been replaced by a single
architecture called the \textbf{transformer,} whimsically named after a
brand of 1980s robot toys, proposed by Vaswani et al.
(\protect\hyperlink{ref-Vaswani}{2017}).

The transformer, a type of neural network, was introduced by researchers
at Google for machine translation tasks. Though we won't go into detail
about how it works, its design was inspired by earlier developments in
neural networks, and it was primarily optimized to allow the GPU-based
simultaneous processing of all parts of even long input texts instead of
word-by-word processing. Earlier architectures were largely
abandoned\footnote{They were not totally abandoned, however, and are
  still used occasionally when datasets are small.} because they didn't
effectively use GPUs and could not process large datasets as quickly.

It didn't take long for researchers to realize that with the transformer
would allow for training models more quickly and/or on more data, as
well as training much \emph{larger models} than other architectures ever
allowed. By ``larger models,'' we mean models with more parameters.
These three elements---larger datasets, faster hardware, and larger
models---all depend on each other. For example, a larger model could
better encode patterns in the training data, but without faster
hardware, training such a model may be infeasible. And if the model is
trained on an insufficient sample of data, it may not generalize
well.\footnote{At its extreme, this phenomenon, known as
  ``overfitting,'' leads to models that ``memorize'' what they see in
  the training data but perform poorly on new data, e.g., the test data.}
Conversely, a substantial dataset may require a larger model (more
parameters) to encode the larger set of discoverable patterns in the
data. Indeed, there is a fundamental tradeoff when selecting
architectures: too few parameters, and the architecture will be limited
in what input-output mappings it can learn, no matter how much training
data is used. Too many parameters (i.e., too large a model), and the
model might overfit.

The simultaneous, rapid increase in datasets and parameter counts, aided
by improved hardware, affected computer vision before affecting NLP. In
fact, the term ``deep learning'' was originally a reference to these
larger models (``deep'' refers to models with increasing numbers of
``layers'' in the architecture, where layers are iterations of repeated
calculations with different parameters at each round). The ``deepening''
of transformers applied to the language modeling task led to what are
now called ``large language models.'' ``Large'' usually refers to the
parameter count, but it could also refer to the size of the training
dataset.

The models in wide use for NLP today have billions of parameters; older
generations of OpenAI models increased from sizes of over a billion
parameters with the largest version of GPT-2 to 175 billion parameters
with GPT-3. The main drawback is that running their training algorithms
on large datasets requires very many GPUs working in parallel for a long
time, which in turn requires a lot of energy. From the perspective of
improving the quality of generated text (in perplexity but also
subjective human judgments), these LLMs represent a major advance.

From a scientific perspective, it's difficult to assess which of these
changes---data size, number of parameters, architecture, etc.---matter
the most. Larger models are more data-hungry; over the last few years,
models have gone from training on datasets with millions of words to
trillions of words. While some work, such as that by Hoffmann et al.
(\protect\hyperlink{ref-chinchilla}{2022}), tried to disentangle the
impacts of model scale and data scale, the additional influence of yet
other factors (like hyperparameters on a training run) complicates
efforts to confidently draw conclusions from such research. These
experiments require the repeated training of models that are estimated
to cost millions of dollars apiece. In addition, it would take far too
long to train fairly matched models based on previously popular,
pre-transformer architectures (i.e., with similar parameter counts on
similar amounts of data to the strongest models of today); this means
that it's impossible to measure how much benefit the transformer offers
other than allowing for larger models.

\noindent\fcolorbox{black}{lightyellow}{%
    \noindent\parbox{\linewidth - 2\fboxsep}{%
        \textbf{It's important to recognize that larger datasets and more powerful hardware were the drivers for the scaling up of language models to architectures with hundreds of billions of parameters (at this writing), and that the parameter count is only part of the explanation for the impressive behaviors of these models.}
    }%
}

\hypertarget{impacts-of-these-changes}{%
\subsection{Impacts of these changes}\label{impacts-of-these-changes}}

What was the impact of LLMs? In short, they caused language modeling
performance to improve dramatically. To see this qualitatively for
yourself, try typing out the beginning of a sentence and instruct a
language model like ChatGPT to complete that sentence. Chances are, you
will immediately see a sentence that reads much more naturally than you
saw generated by a simpler autocomplete system at the beginning of this
section. Many people have shared this subjective experience of more
fluent text generation, and it is backed up by quantitative evaluations
like perplexity. However, if that were their only contribution, LLMs
probably wouldn't have entered the public consciousness.

\hypertarget{alllmsnow}{%
\subsubsection{Many other tasks are now reduced to language
modeling}\label{alllmsnow}}

We previously mentioned in section \ref{whylm} that LMs could inform NLP
systems designed for \emph{other} tasks. LLMs are accelerating this
trend. By formulating task instructions in natural language, perhaps
also providing additional specific examples of what it would look like
to successfully perform the task (inputs and outputs), and then
supplying that text as the context on which a LLM conditions when
choosing next words as continuations, we see very reasonable outputs for
a broad range of such tasks (e.g., generating summaries and answering
questions). As we discussed in section \ref{whylm}, many techniques
built on the pretraining-finetuning approach transferred strong language
model performance to other tasks. But the extent to which LLMs became
the \emph{full} model pipeline, i.e., with no task-specific finetuning
needed for particular tasks, was striking.\footnote{The idea of
  prompting a model with a small number of examples came to be known as
  ``in-context learning.'' Considerable effort has gone into engineering
  prompts for better task performance and into finetuning LMs to follow
  instructions describing widely varied tasks. Such instruction
  finetuning has become a widely used second stage of training for
  commercial LM products. Note that it requires a dataset of
  instructions paired with the desired response an LM should give to
  each.} Importantly, remember that part of the definition of a task is
an evaluation method; the striking observation is that, as language
models achieve lower perplexity, they also achieve better performance on
many other tasks' own evaluations.

For example, we previously described translation between languages and
sentiment analysis as two broad categories of NLP applications. Today's
LLMs can often perform those tasks given context instructions and/or
examples --- i.e., they are ``prompted'' to do so. For example, consider
a context like ``Translate this sentence into French: We'd like another
bottle of wine.'' If an LLM has seen enough text that includes
requests/responses, text in the relevant languages, and parallel
examples, it could produce the translation. (Indeed, OpenAI's ChatGPT
system gave us a fairly reasonable ``Nous aimerions une autre bouteille
de vin.'' Similarly, the prompt ``Is the sentiment toward the movie
positive or negative? This film made me laugh, but only because it was
so poorly executed.'' led ChatGPT to output that the sentiment was
negative.)

This ease of transferability has made it much simpler for a wider
variety of people, including non-researchers, to explore NLP
capabilities. Often, it is no longer necessary to collect training data
and build a specialized model for a task. We can say what we want in
natural language to prompt an LLM, and we will often get output close to
what we intended. People, including experts and non-experts, are now
using LLMs for many purposes, including many not originally formalized
as NLP tasks.

\hypertarget{black-boxes}{%
\subsubsection{Black boxes}\label{black-boxes}}

Modern transformers are considered to be ``black boxes'' with befuddling
numbers of parameter-knobs to turn, and to our knowledge, no one has
particularly useful intuition about how to set any particular knob. This
situation seems daunting, like sitting in a cockpit with thousands of
knobs and controls and being told to fly the plane with no training.
Indeed, it's only because of the increasing computational power of
commercially available computers that we can solve problems this way
today, but this still leaves us without a sense of the kinds of
information models have learned to leverage, or how.

Both the transformer architecture and the stochastic gradient descent
method used to set its parameters are mystifying, at least at first.
Below, we reflect on that and note important differences that make an
architecture like the transformer more inscrutable.

Stochastic gradient descent, the algorithm used to train transformers
and other neural networks, has been extensively studied and is very well
understood for some kinds of problems. Picture a smooth bowl and imagine
a marble placed anywhere in it. That marble will roll and eventually
settle at the lowest point. If the dish were sitting on a piece of graph
paper (a two-dimensional plane), the coordinates of that lowest point
are the values of our two parameters that minimize the loss function.
Stochastic gradient descent is, roughly speaking, doing the work of
gravity. The simple curve of the dish, with no bumps or cutouts or
chips, corresponds to the property of convexity. Some machine learning
problems correspond to a convex loss function, and theoretical proofs
support the existence of the best parameter values, how close SGD gets
to them, and how fast. What remains surprising is that SGD works well in
practice even when the loss function is not convex (like the Cascades,
discussed in section \ref{minimization}). But the mathematics underlying
this algorithm are relatively mature.

The transformer architecture, only a few years old at this writing,
remains mysterious. Some researchers have sought to prove theorems about
its limitations (i.e., input-output mappings it cannot represent under
some conditions), and more have run experiments to try to characterize
what it learns from data in practice. More research is clearly needed,
both to improve our understanding of what we can expect from this
architecture and to help define new architectures that work better or
for which parameter setting is less computationally expensive.

\hypertarget{cost-and-complexity-affect-who-can-develop-these-models-now}{%
\subsubsection{Cost and complexity affect who can develop these models
now}\label{cost-and-complexity-affect-who-can-develop-these-models-now}}

Yet another effect of the move to LLMs has been that a much smaller set
of organizations can afford to produce such models. Since large,
well-funded tech companies are (almost exclusively) well positioned to
train LLMs due to their access to both data and specialized hardware,
these companies are the sources for almost all current LLMs. This poses
a barrier to entry for many researchers at other institutions. Given the
wide array of different communities that could benefit from using these
models, the many different purposes they might envision for these
models, and the vast diversity of language varieties that they
represent, determining ways to broaden participation in LLM development
is an important emerging challenge.

Furthermore, when models were smaller, the idea of ``running out'' of
web text on the public internet seemed ludicrous; now, that's a looming
concern for LLM developers. As massive datasets play an increasingly
large role in model training, some large companies' access to their own
massive proprietary data associated with platforms they maintain may
give them an advantage in their development of models of text.

\hypertarget{adaptingasproducts}{%
\subsubsection{Adapting LLMs for use as
products}\label{adaptingasproducts}}

Because of the capabilities of these new models, many developers seek to
integrate them into a wide array of products and services, from helping
software engineers write code to helping lawyers write briefs. This
echoes a longstanding practice of incorporating LMs into parts of
standalone products with commercial purposes, such as guiding a
translation system to produce more fluent text in the output language.
As LLMs gained broader exposure (and, we conjecture, with increased
internal testing at the companies where they were built), it became
clear that additional adjustments were needed before deploying these
models in products.

We relate some of the more concerning issues that emerge in
LLM-generated text in section \ref{capabilities}. For now, consider the
concrete possibility that an LLM would generate text that is fluent, but
impolite or even obscene. How can this be prevented? Enforcing
conventions of social acceptability is a difficult problem that many
researchers have tackled. Proposed methods can vary from post-processing
outputs (e.g., to screen out outputs that include certain dispreferred
words) to reranking sampled outputs using an auxiliary model
specifically trained on curated data to exhibit politeness. It is
difficult to ``taskify'' social acceptability because it is
context-dependent and extremely subjective.

The notion of ``alignment,'' often used today for this class of
problems, was introduced by Norbert Wiener: ``If we use, to achieve our
purposes, a mechanical agency with whose operation we cannot efficiently
interfere\ldots{} then we had better be quite sure that the purpose put
into the machine is the purpose which we really desire''
(\protect\hyperlink{ref-norbertwiener}{Wiener 1960}). This idea comes
through today in research on using machine learning to alter LM
behaviors directly.

In practice, commercial models are further trained on tasks designed to
encourage instruction following (section \ref{alllmsnow}) and generating
text that humans respond to favorably.\footnote{One current example of a
  proposed method for doing this is ``reinforcement learning from human
  feedback.'' As its name implies, this method uses machine learning to
  turn discrete representations of human preferences, like ``sampled
  output A is preferable to sampled output B,'' into a signal for how to
  adjust a model's parameters accordingly.} It is complicated to
determine \emph{which} behaviors to encourage. In her 2023 keynote at
the FAccT research conference, the social scientist Alondra Nelson made
the point that ``civilizations, for eons, for millennia\ldots{} choose
your long time scale---have been struggling, fighting, arguing, debating
over human values and principles''
(\protect\hyperlink{ref-alondranelson}{Nelson 2023}). In other words,
not only is it a difficult problem to determine how to shape models'
outputs to reflect a given set of values, it's also extremely
complicated to determine which set of values to incorporate into that
set. Therefore, we tend to view these last adjustments of an LLM's
behavior as a kind of customization rather than as an intrinsic encoding
of ``human values'' into the system. Just like training models, only a
few companies are currently equipped to customize them at this writing.

\hypertarget{safeguards}{%
\subsubsection{Safeguards and mitigation}\label{safeguards}}

Because LLMs are trained on such a wide-variety of internet content,
models can create outputs that contain unsafe content. For example, a
user may want to know how to create a bomb or have the model help them
plan some other dangerous or illegal act. Leaving aside whether the
models constitute ``intelligence,'' the information these models contain
and how easily they present it to users can create substantial risk. The
current method for attempting to solve this problem is establishing
\textbf{content safeguards,} a major part of adapting LLMs for use as
products. Safeguards can take different forms, from tuning the model to
avoid certain topics to addressing the issue through post-processing,
where output from the model is filtered. These safeguards are part of
the larger ``alignment'' process since they can also be used to help
block hateful content in addition to dangerous information.

There are also less obvious cases where safeguards can be critical for
user safety. For example, a model should not provide medical advice
without at least suggesting that the user seek professional advice and
disclosing that it is not a doctor or that its output is not guaranteed
to be consistent with the medical community's consensus. Another case is
self-harm, where the behavior of LLMs has been likened to a mirror,
e.g., encouraging behaviors reflected in user prompts.

Though necessary, safeguards can also impact a model's utility depending
on how they are implemented. For example, a model that is too strict may
refuse to do something that isn't actually harmful, making it less
useful. Therefore, there is a tension between cautiously avoiding
liability for model developers and meeting user expectations.

\hypertarget{evaluationcrisis}{%
\subsubsection{The evaluation crisis}\label{evaluationcrisis}}

Excitement around LLMs often centers on the rate of progress: as the
models get larger (or are trained on more data), they seem to get
increasingly accurate and fluent. As mentioned previously in section
\ref{taskification}, NLP researchers have long-standing, rigorous
methods for measuring how well systems perform at various tasks. These
have not been abandoned. Following the trend of adapting LLMs to almost
every task NLP originally set out to do, with relatively little transfer
effort (section \ref{alllmsnow}), researchers are now evaluating new
models, adapted in new ways, on ever-growing suites of tasks drawn from
the past few decades of empirical evaluation of NLP systems, as well as
new ones coming into use. The general trend is that performance numbers
are improving.

This is promising news insofar as these tasks accurately capture what
people want to do with NLP technology. But we believe there are reasons
to be skeptical. Since the deployment and widespread adoption of
LLM-based products, users have expressed enthusiastic interest in
thousands of new use cases for LLMs that bear little resemblance to the
tasks that constitute our standard research evaluations, which has
several important implications:

\begin{itemize}
\tightlist
\item
  The suite of tasks driving research evaluations needs thorough and
  ongoing reconsideration and updating to focus on communities of actual
  users.
\item
  Observations of how real users interact with an LLM, along with
  feedback on the quality of the LLM's behavior, will be important for
  continuing to improve LLM quality.
\item
  Because there is diversity in the communities of users, customization
  of models will become increasingly important, making thorough
  evaluation increasingly multi-faceted and challenging.
\item
  Reports of ``progress'' cannot be taken at face value; there are many
  different aspects to model quality. A single performance number (like
  perplexity on a test set or average performance on a suite of hundreds
  or thousands of tasks' specific evaluations) will not meaningfully
  convey the strengths and weaknesses of a system with such wide-ranging
  possible behaviors.
\end{itemize}

We believe that these challenges will inspire new collaborations between
researchers and users to define evaluations (and, by extension, models)
that work as our needs and the realities of model building evolve..

\hypertarget{knowingitstrainingdata}{%
\subsection{Knowing the model means knowing its training
data}\label{knowingitstrainingdata}}

Model capabilities depend directly on the specific data used to train
them. The closer a string of text (say, the instructions provided to an
LLM) is to the kind of data that the model was trained on (which, for
current models, is a large portion of the data on the internet), the
better we expect that model to do in mimicking reasonable continuations
of that ``kind'' of language.\footnote{Note that we are \emph{not}
  implying that language models are only mimics; characterizing the
  precise ways in which they merely copy vs.~generalize is work still to
  be done.} Conversely, the further the language of some text is from
the model's training data, the less predictable the model's continuation
of that text will be. (In section \ref{wording}, we discuss the
implications for choosing which prompts to supply to a model.)

You can test this out. Try instructing a model (for example, ChatGPT) to
generate some text (a public awareness statement, perhaps, or a plan for
an advertising campaign) about a very specific item X geared towards a
specific subpopulation Y, preferably with an X and Y that haven't
famously been paired together.

Grammatically, the answer returned is probably fine. However, if the
content of the model's response seems generic, that's not too
surprising. The amount of text that models like ChatGPT are trained on
that could serve as a close example to a particular prompt is typically
\emph{far} greater than that which is relevant for precise ideas
specific to whatever personal combination you thought up.

If you speak a language besides English, you'll likely also notice a
worse answer or a more stilted, generic tone if you translate your
question into that language and ask it again. And again, this is
directly related to the model's training data: however much text there
is relevant to your issue or product on the internet in English, there's
likely less of it in your other language, meaning there was less
available to use for training.

\hypertarget{trainingdatacontents}{%
\subsubsection{What does LLMs' training data
contain?}\label{trainingdatacontents}}

Characterizing a dataset on a trillions-of-words scale is tricky for a
few reasons. First, reading through the corpus, or even a large enough
sample to capture its diversity, would take too long. (A colleague of
ours estimated thousands of years of reading without any breaks.)
Published descriptions of datasets that have been explored using
automated tools focus on the top sources (e.g., web domains like
Reddit.com or Wikipedia.org) or coarse characterizations in terms of
genre (e.g., patents, news); see Dodge et al.
(\protect\hyperlink{ref-dodge-etal-2021-documenting}{2021}) for an
example. These characterizations, while convenient, show tremendous
variation. We believe that researchers must do more work on developing
methodologies and implementing tools for describing that variation.

In many cases, though, information about the documents used to train an
LLM is hidden. It's very common for companies that deploy these models
to treat the data they used as a trade secret, saying little to nothing
about the data, making analysis impossible. However, a few model
builders do share more information about their training data, which
helps researchers better understand how model behaviors, beneficial and
otherwise, are shaped by certain kinds of text.

Many researchers have one specific concern about hidden training
datasets: Suppose a model is prompted with a question that seems
especially difficult to answer, and it answers accurately and clearly,
like an expert. We should be impressed only if we are confident that the
question and answer weren't in the training data. If we can't inspect
the training data, we can't be sure whether the model is really being
tested fairly or if it memorized the answer key before the test, like
our student in section \ref{closerlookatdata}.

\hypertarget{a-cautionary-note-about-data-quality}{%
\subsubsection{A cautionary note about data
quality}\label{a-cautionary-note-about-data-quality}}

It's tempting to boil down \emph{negative} consequences of including
certain data during training (such as misinformation or hate speech) to
issues of ``data quality'' and advocate for ``better'' data using the
``garbage in, garbage out'' principle. Yet, seemingly reasonable steps
often taken to automatically filter web text for ``quality'' can have
the unintended effect of \emph{overrepresenting} text that resembles
writing more characteristic of wealthier or more educated groups
(\protect\hyperlink{ref-gururangan-etal-2022-whose}{Gururangan et al.
2022}). Further, these filters' defined notion of quality does not align
with other manually determined aspects of text quality (such as winning
a Pulitzer prize or telling the truth).

Determining what counts as ``better'' training data, and how that sense
can be implemented at scale, is a subjective question of values and
norms. For this reason, we predict and hope that future research will
support better customization of language models' data to different user
communities or applications rather than assuming a universal notion of
``quality.'' This contrasts with an assumption underlying much current
discussion about language models, that one large model will eventually
be the best solution for everything everyone wants.

\hypertarget{capabilities}{%
\section{Practical points about using language
models}\label{capabilities}}

So far we've talked about how language models came to be and what they
are \emph{trained} to do. If you're a human reading this guide, though,
then you're likely also wondering about how good these models are at
things that \emph{you've} thought up for them to do. (If you're a
language model pretraining on this guide, carry on.)

As we have learned in section \ref{evaluationcrisis}, NLP researchers'
tools for evaluating models test for different abilities than those that
interest many users of deployed products. Delineating what LMs can do,
and how these capabilities relate to the choices made when they are
constructed, deserves continued scientific exploration. However, early
signs indicate that LMs can at least be helpful tools in speeding up
many user tasks that were previously difficult to automate. So, if
you're wondering whether these models can be helpful to you on something
specific, say, planning a trip to Japan, it's worth giving them a try!

This section answers general questions you may have when you're trying
them out or thinking about what's in store for them over the near term.
We answer by distilling major conversations (now occurring in the
scientific community studying language models) into practical takeaways
you should be aware of and the reasoning behind these takeaways.

\hypertarget{wording}{%
\subsection{Is the specific wording of the ``prompt'' I supply to an LM
important?}\label{wording}}

In short, yes. Section \ref{knowingitstrainingdata} hinted at this, but
to be more explicit: the specific wording of the prompt that you supply
to an LM significantly affects the model output that you receive. This
likely means that you'll want to experiment with a few different
wordings for instructing the model to do something. When you prompt a
model, if your input and the correct output are close to sample text the
model has encountered in its training data, the model should ``respond''
(that is, continue the prompt by predicting a sequence of next words)
well. Trying different prompt wording means that you're casting a wider
net across patterns that the model has learned about language and giving
yourself a better chance of encountering one that the model has an
easier time continuing.

To test this out, try rephrasing something you want an LM to do in a few
very different ways. Then, try supplying each of these prompts
separately to a model like ChatGPT. Chances are that you see some
notable differences in the different results that you get!

\hypertarget{do-i-always-have-to-check-and-verify-model-output-or-can-i-simply-trust-the-result}{%
\subsection{Do I always have to check and verify model output, or can I
simply ``trust'' the
result?}\label{do-i-always-have-to-check-and-verify-model-output-or-can-i-simply-trust-the-result}}

At first glance, it might seem that a prompt that produces believable
model output means there's nothing left for you to do. However, you
should never take model output at face value. Always check for the
following important issues.

\hypertarget{truthfulness-vs.-hallucination}{%
\subsubsection{Truthfulness
vs.~``hallucination''}\label{truthfulness-vs.-hallucination}}

At the time of writing (and likely for the foreseeable future), LMs
struggle with `telling the truth,' that is, producing correct output. In
fact, a much-discussed property of LMs is their tendency to produce
inaccurate and nonfactual information. This phenomenon is known as
\textbf{``hallucination.''}\footnote{Some have argued that the term
  ``hallucination'' is misleading and anthropomorphizes language models,
  but at this writing it is the most widely used by NLP researchers.}
How much hallucination matters greatly depends on the tasks and genres
of language of the model's users. For a creative writer, a language
model's flexibility in presenting fictional information may be one of
its greatest strengths. For someone who needs an accurate summary of a
medical article or who tries to use an LM to retrieve statements of fact
from court testimony, it can render the model unusable, at least without
careful post-prompt fact-checking.

Why do models hallucinate? While models depend heavily on their training
data, they do not access that data \emph{exactly.} Instead, they seem to
encode patterns in the data, but not to ``remember'' the data precisely
all the time. Thus, for topics with plenty of supporting data and a
simple task, the likelihood of hallucination is often lower. With more
complex tasks on less-discussed subjects, hallucination is less
surprising. Even when there is plenty of data, if the training data
included frequent statements of incorrect information (for example, the
incorrect but widely discussed claim that vaccines cause autism), the
model may encode (as a pattern) the incorrect claim. There is ongoing
active research on discouraging models from stating incorrect
information as well as steering them away from generating
confident-sounding answers (or any answer at all) to questions where the
facts may be under debate, but this is still a very difficult open
problem.

Relatedly, there is currently no straightforward, computationally
feasible way to link specific predictions or generated text back to
specific training documents or paragraphs. So, another ongoing research
challenge is endowing LMs with the ability to ``cite their sources,''
that is, to not only generate explicit and accurate references to
relevant literature or sources like scholars are taught to do, but to
reveal the specific texts that influenced a particular next word
prediction, if requested.

A notable real-life example of these missing capabilities surfaced when
two US lawyers in early 2023 used ChatGPT to prepare the filing for a
personal injury suit against an airline. While the main text was very
fluent, the model had completely hallucinated the cases it cited as
precedents and their corresponding judges, plaintiffs, and defendants.
This was brought to the court's attention when it received a brief from
the airline's lawyers questioning the existence of the cited cases.
These cases weren't real, and the lawyers had not disclosed that they
used ChatGPT for their legal research. The federal judge in the case was
furious and fined both lawyers, who blamed ChatGPT during a subsequent
hearing, stating they ``did not understand it was not a search engine,
but a generative language-processing tool.''

Now that LM hallucinations have found their way into the judicial
system, we can hope that users (and model builders, the ``deep pockets''
in such cases) have learned a lesson. LMs are \emph{not} search engines,
and their output requires careful checking, at least at present.

\noindent\fcolorbox{black}{lightyellow}{%
    \noindent\parbox{\linewidth - 2\fboxsep}{%
        \textbf{Remember:  language models don't perfectly capture their training data!}
    }%
}

\hypertarget{biases}{%
\subsubsection{Model outputs that reflect social biases}\label{biases}}

Another aspect of evaluating and revising model outputs where human
judgment is key is in checking for models' unthinking mimicry of social
biases that may have appeared in their training data.

NLP researchers often refer to the names of the idealized tasks we've
trained our models to perform---``hate speech detection,'' ``machine
translation,'' ``language modeling''---but remember that how a model
learns to perform a task is heavily influenced by the particular data
used to train it. (This is related to our previous discussion in section
\ref{taskification} about the tradeoff between abstract, aspirational
notions of a task and concrete, workable ones.) In practice, models for
``hate speech detection'' are actually trained to perform ``hate speech
detection as exemplified in the HateXplain dataset'' or ``hate speech
detection as exemplified in the IberEval 2018 dataset.'' These datasets
reflect their builders' focus on particular type(s) of language---for
example, Spanish-language news articles or American teenagers' social
media posts---but no dataset perfectly represents the type(s) of
language it's meant to represent. There are simply too many possible
utterances! Therefore, despite ongoing work trying to improve models'
abilities to generalize from the data observed during training, it
remains possible that a model will learn a version of the task that's
informed by quirks of its training data. Because there are so many
possible ``quirks,'' it's a safe bet that a model \emph{will} have
learned some of them. And in fact, we've observed this time and again in
NLP systems.

To be more specific, let's look at some past work that's found bias
traceable to the training data within hate speech detection systems. Sap
et al. (\protect\hyperlink{ref-sap-etal-2019-risk}{2019}) found that in
two separate hate speech detection datasets, tweets written in African
American Vernacular English (AAVE) were disproportionately more likely
to be labeled as toxic than those written in white-aligned English by
the humans employed to detect toxicity. Not only that, but models
trained on those datasets were then more likely to \emph{mistakenly}
label innocuous AAVE language as toxic than they were to mistakenly flag
innocuous tweets in white-aligned English. This gives us an idea of how
dataset bias can propagate to models in text classification systems, but
what about in cases where models generate text? If models aren't
associating text with any human-assigned toxicity labels, how can they
demonstrate bias?

As it turns out, evidence of bias is still visible even in cases where
the model isn't generating a single predefined category for a piece of
text. A famous early example of work showing this for Google Translate
based its study on a variety of occupations for which the US Bureau of
Labor Statistics publishes gender ratios
(\protect\hyperlink{ref-Prates2019}{Prates, Avelar, and Lamb 2019}). The
authors evaluated machine translation systems that translated to English
from various languages that don't use gendered singular pronouns,
constructing sentences such as ``{[}neutral pronoun{]} is an engineer''
and translating them into English. They found that these systems
demonstrated a preference for translating to ``he'' that often far
exceeded the actual degree by which men outnumbered women in an
occupation in the US. This bias likely reflects an imbalance in the
number of training sentences associating men and women with these
different professions, indicating another way in which a skew in the
training data for a task can influence a model.

Imbalances like this are examples of those ``quirks'' we mentioned
earlier, and they can be puzzling. Some quirks, like data containing far
more mentions of male politicians than female politicians, seem to
follow from the prevalence of those two categories in the real world.
Other quirks initially seem to defy common sense: though black sheep are
not prevalent in the world, ``black sheep'' get mentioned more often in
English text than other-colored sheep, perhaps because they're more
surprising and worthy of mention (or perhaps because a common idiom,
``the black sheep of the family,'' uses the phrase).

In the same way that biases can arise in machine translation systems,
LMs can exhibit bias in generating text. While current LMs are trained
on a large portion of the internet, text on the internet can still
exhibit biases that might be spurious and purely accidental, or that
might be associated with all kinds of underlying factors: cultural,
social, racial, age, gender, political, etc. Very quickly, the risks
associated with deploying real-world systems become apparent if these
biases are not checked. Machine learning systems have already been
deployed by private and government organizations to automate high-stakes
decisions, like hiring and determining eligibility for parole, which
have been shown to discriminate based on such factors
(\protect\hyperlink{ref-raghavan-hiring}{Raghavan et al. 2020};
\protect\hyperlink{ref-nishi2019privatizing}{Nishi 2019}).

So how exactly can researchers prevent models from exhibiting these
biases and having these effects? It's not a solved problem yet, and some
NLP researchers would argue that these technologies simply shouldn't be
used for these types of systems, at least until there is a reliable
solution. For LMs deployed for general use, research is ongoing into
ways to make models less likely to exhibit certain known forms of bias
(e.g., see section \ref{adaptingasproducts}). Progress on such research
depends on iterative improvements to data and evaluations that let
researchers quantitatively and reproducibly measure the various forms of
bias we want to remove.

\noindent\fcolorbox{black}{lightyellow}{%
    \noindent\parbox{\linewidth - 2\fboxsep}{%
        \textbf{Remember:  datasets and evaluations never perfectly capture the ideal task!}
    }%
}

\hypertarget{are-language-models-intelligent}{%
\subsection{Are language models
intelligent?}\label{are-language-models-intelligent}}

The emergence of language model products has fueled many conversations,
including some that question whether these models might represent a form
of ``intelligence.'' In particular, some have questioned whether we have
already begun to develop ``artificial general intelligence'' (AGI). This
idea implies something much bigger than an ability to do tasks with
language. What do these discussions imply for potential users of these
models?

We believe that these discussions are largely \emph{separate} from
practical concerns. Until now in this document, we've mostly chosen used
the term ``natural language processing'' instead of ``artificial
intelligence.'' In part, we have made this choice to scope discussion
around technologies for language specifically. However, as language
model products are increasingly used in tandem with models of other
kinds of data (e.g., images, programming language code, and more), and
given access to external software systems (e.g., web search), it's
becoming clear that language models are being used for more than just
producing fluent text. In fact, much of the discussion about these
systems tends to refer to them as examples of AI (or to refer to
individual systems as ``AIs'').

A difficulty with the term ``AI'' is its lack of a clear definition.
Most uncontroversially, it functions as a descriptor of several
different \emph{communities} researching or developing systems that, in
an unspecified sense, behave ``intelligently.'' Exactly what we consider
intelligent behavior for a system shifts over time as society becomes
familiar with techniques. Early computers did arithmetic calculations
faster than humans, but were they ``intelligent?'' And the applications
on ``smart'' phones (at their best) don't seem as ``intelligent'' to
people who grew up with those capabilities as they did to their first
users.

But there's a deeper problem with the term, which is the notion of
``intelligence'' itself. Are the capabilities of humans that we consider
``intelligent'' relevant to the capabilities of existing or hypothetical
``AI'' systems? The variation in human abilities and behaviors, often
used to explain our notions of human intelligence, may be quite
different from the variation we see in machine intelligence. In her 2023
keynote at ACL (one of the main NLP research conferences), the
psychologist Alison Gopnik noted that in cognitive science, it's widely
understood that ``there's no such thing as general intelligence, natural
or artificial,'' but rather many different capabilities that cannot all
be maximally attained by a single agent
(\protect\hyperlink{ref-gopnik}{Gopnik 2023}).

In that same keynote, Gopnik also mentioned that, in her framing,
``cultural technologies'' like language models, writing, or libraries
can be impactful for a society, but it's people's learned use of them
that make them impactful, not inherent ``intelligence'' of the
technology itself. This distinction, we believe, echoes a longstanding
debate in yet another computing research community, human-computer
interaction. There, the debate is framed around the development of
``intelligence augmentation'' tools, which humans directly manipulate
and deeply understand, still taking complete responsibility for their
own actions, vs.~agents, to which humans delegate tasks
(\protect\hyperlink{ref-shneiderman-97}{Shneiderman and Maes 1997}).

Notwithstanding debates among scholars, some companies like OpenAI and
Anthropic state that developing AGI is their ultimate goal. We recommend
first that you recognize that ``AGI'' is not a well-defined scientific
concept; for example, there is no agreed-upon test for whether a system
has attained AGI. The term should therefore be understood as a marketing
device, similar to saying that a detergent makes clothes smell ``fresh''
or that a car is ``luxurious.'' Second, we recommend that you assess
more concrete claims about models' specific \emph{capabilities} using
the tools that NLP researchers have developed for this purpose. You
should expect no product to ``do anything you ask,'' and the clear
demonstration that it has one capability should never be taken as
evidence that it has different or broader capabilities. Third, we
emphasize that AGI is not the explicit or implicit goal of all
researchers or developers of AI systems. In fact, some are far more
excited about tools that \emph{augment} human abilities than about
autonomous agents with abilities that can be compared to those of
humans.

We close with an observation. Until the recent advent of tools marketed
as ``AI,'' our experience with intelligence has been primarily with
other humans, whose intelligence is a bundle of a wide range of
capabilities we take for granted. Language models have, at the very
least, linguistic fluency: the text they generate tends to follow
naturally from their prompts, perhaps indistinguishably well from
humans. But LMs don't have the whole package of intelligence that we
associate with humans. In language models, fluency, for example, seems
to have been separated from the rest of the intelligence bundle we find
in each other. We should expect this phenomenon to be quite shocking
because we haven't seen it before! And indeed, many of the heated
debates around LMs and current AI systems more generally center on this
``unbundled'' intelligence. Are the systems intelligent? Are they more
intelligent than humans? Are they intelligent in the same ways as
humans? If the behaviors are in some ways indistinguishable from human
behaviors, does it matter that they were obtained or are carried out
differently than for humans?

We suspect that these questions will keep philosophers busy for some
time to come. For most of us who work directly with the models or use
them in our daily lives, there are far more pressing questions to ask.
What do I want the language model to do? What do I \emph{not} want it to
do? How successful is it at doing what I want, and how readily can I
discover when it fails or trespasses into proscribed behaviors? We hope
that our discussion helps you devise your own answers to these
questions.

\noindent\fcolorbox{black}{lightyellow}{%
    \noindent\parbox{\linewidth - 2\fboxsep}{%
        \textbf{Remember: analogies to human capabilities never perfectly capture the capabilities of language models, and it's important to explicitly test a model for any specific capability that your use case requires!}
    }%
}

\hypertarget{future}{%
\section{Where is the development of language models
headed?}\label{future}}

Language models (and the role they play in society) are still in their
infancy, and it's too early to say how they will continue to develop and
the main ways in which they will evolve over time. Currently, as we've
mentioned, most language models (and generative AI models more
generally) are developed by a handful of companies that are not very
forthcoming about their construction. However, it's important to
remember that, depending on various factors over the next several years,
a future of more decentralized models managed by not-for-profit entities
is still possible.

One key variable that's still taking shape in determining this future is
governed by democratic processes: government regulation, in the form of
policy and law. This means that public attention (your attention) to
issues around these models could directly influence what the future of
the technology looks like. We now discuss both the reasons for
difficulties in predicting the future of language model development and
the role that early regulation of these models has played so far.

\hypertarget{why-is-it-difficult-to-make-projections-about-the-future-of-nlp-technologies}{%
\subsection{Why is it difficult to make projections about the future of
NLP
technologies?}\label{why-is-it-difficult-to-make-projections-about-the-future-of-nlp-technologies}}

For perspective, let's consider two past shifts in the field of NLP that
happened over the last ten years. The first, in the early 2010s, was a
shift from statistical methods---where each parameter fulfilled a
specific, understandable (to experts) role in a probabilistic model---to
neural networks, where blocks of parameters without a corresponding
interpretation were learned via gradient descent. The second shift,
around 2018--19, was the general adoption of the transformer
architecture we described in section \ref{transformers}, which mostly
replaced past neural network architectures popular within NLP, and the
rise of language model pretraining (as discussed in section
\ref{whylm}).

Most in the field didn't anticipate either of those changes, and both
faced skepticism. In the 2000s, neural networks were still largely an
idea on the margins of NLP that hadn't yet demonstrated practical use;
further, prior to the introduction of the transformer, another, very
different structure of neural network\footnote{It was called the LSTM,
  ``long short-term memory'' network.} was ubiquitous in NLP research,
with relatively little discussion about replacing it. Indeed, for
longtime observers of NLP, one of the few seeming certainties is of a
significant shift in the field every few years---whether in the form of
problems studied, resources used, or strategies for developing models.
The form this shift takes does not necessarily follow from the dominant
themes of the field over the preceding years, making it more
``revolutionary'' than ``evolutionary.'' And, as more researchers are
entering NLP and more diverse groups collaborate to consider which
methods or which applications to focus on next, predicting the direction
of these changes becomes even more daunting.

A similar difficulty applies when thinking about long-term real-world
\emph{impacts} of NLP technologies. Even setting aside that we don't
know how NLP technology will develop, determining how a particular
technology will be used poses a difficult societal question.
Furthermore, NLP systems are being far more widely deployed in
commercial applications; this means that model developers are getting
far more feedback about them from a wider range of users, but we don't
yet know the effects that deployment and popular attention will have on
the field.

Remembering how these models work at a fundamental level---using
preceding context to predict the next text, word by word, based on what
worked best to mimic demonstrations observed during training---and
imagining the kinds of use cases that textual mimicry is best-suited
towards will help us all stay grounded and make sense of new
developments.

\hypertarget{what-might-ai-regulation-look-like}{%
\subsection{What might AI regulation look
like?}\label{what-might-ai-regulation-look-like}}

An important conversation about the future of language models centers
around possible regulation of these models. This topic encompasses many
related discussions: companies' self-regulation, auditing of models by
third parties, restrictions on data collection by private companies
(such as those recently instituted by Reddit), and potential government
oversight. Given that companies producing these models must already make
decisions about how to adjust their models' behavior, it seems most
realistic to consider not \emph{whether} regulation by some party will
occur, but rather \emph{which} forms of regulation would be beneficial.
We will first describe some early attempts at regulating AI and then
hypothesize about what future regulations might focus on.

Before doing that, we make one additional point. It's worth bearing in
mind that calls in the public sphere for or against regulation can arise
for a variety of different reasons. For example, as Kevin Roose
\href{https://www.nytimes.com/2023/07/11/technology/anthropic-ai-claude-chatbot.html}{recently
wrote} for the \emph{New York Times}, ``some skeptics have suggested
that A.I. labs are stoking fear out of self-interest, or hyping up
A.I.'s destructive potential as a kind of backdoor marketing tactic for
their own products. (After all, who wouldn't be tempted to use a chatbot
so powerful that it might wipe out humanity?)''\footnote{See also
  \href{https://www.nytimes.com/2023/09/28/opinion/ai-safety-ethics-effective.html}{this
  opinion article} by Bruce Schneier and Nathan Sanders.} Past a certain
point, discussion of AI regulation can become politically charged,
drawing on many complicated variations of societal values. Therefore,
similar to when participating in any public discussion, it's helpful to
get in the habit of thinking about \emph{why} a specific person might be
saying what they're saying given their background and interests, as well
as \emph{who} they're hoping their comments will influence.

\hypertarget{emergingregulation}{%
\subsubsection{What versions of government AI regulation are
emerging?}\label{emergingregulation}}

In terms of concrete regulation that has made its way into the sphere of
public policy, US President Joe Biden's
\href{https://www.whitehouse.gov/briefing-room/presidential-actions/2023/10/30/executive-order-on-the-safe-secure-and-trustworthy-development-and-use-of-artificial-intelligence/}{Executive
Order on AI} and the European Union's
\href{https://artificialintelligenceact.eu/}{2023 AI Act} represent the
most sweeping regulatory measures relating to AI thus far.

The Executive Order on AI, made at the end of October 2023, set out to
establish general principles around AI innovation. These were high-level
and focused primarily on the management of AI risk and security, the
promotion of responsible AI innovation and competition, and the
protection of individuals and their civil liberties as AI continues to
advance. An additional focus of the order is to garner AI talent in the
United States and the US government. While these points are focused on
the promotion of AI, the order also includes a threshold of required
computing power where a model could be used in ``malicious cyber-enabled
activity.'' That is, if a specific number of floating-point operations
used in the training of a model is exceeded, then some uses of that
model might be considered a risk. This definition reflects the
difficulty of translating the high-level concept of ``model risk'' into
lower-level terms; it is quite possible that there will be further
iterations of this definition in response to the continued advancement
of computing capabilities.

The focus of the EU AI Act is the determination of a risk level posed by
different AI systems to human individuals based on proposed and likely
use cases of those systems, for the purposes of identifying higher-risk
technologies and restricting their use. The details of the AI Act are
also fairly high-level and ultimately most of the act was effectively
upended by the sudden widespread surge in use of ChatGPT. The AI Act was
a lodestone for political debates over the extent to which AI regulation
should affect different systems, with positions influenced by concerns
as varied as fostering support for scientific innovation or upholding
the rights of those affected by model decisions. The upending of the EU
AI Act shows that whatever future regulation is released likely won't
regulate for a certain point in time---as we are already seeing in some
ways with the Executive Order on AI. Any regulation that isn't focused
on broader concepts like harm reduction and safe use cases runs the risk
of becoming quickly outdated, given the current (and likely future) pace
of technology development.

At a lower level closer to the implementation and training of AI
systems, the legal focus so far has overwhelmingly been on copyrights
associated with models' training data. A
\href{https://www.lexology.com/library/detail.aspx?g=68d490a1-3021-4040-afdd-90ae8fa69337}{2018
amendment to Japan's 1970 Copyright Act} gives generative AI models
broad leeway to train on copyrighted materials provided that the
training's ``purpose is not to enjoy the ideas or sentiments expressed
in the work for oneself or to have others enjoy the work.'' However,
more recent court cases focused on generative image models, such as
\href{https://www.reuters.com/legal/getty-images-lawsuit-says-stability-ai-misused-photos-train-ai-2023-02-06/}{Getty
Images suing Stability AI Inc.} or
\href{https://news.bloomberglaw.com/ip-law/ai-art-generators-hit-with-copyright-suit-over-artists-images}{a
group of artists suing Stability AI, Midjourney, and DeviantArt}, are
pushing back on that view and have yet to reach a resolution.

Even these early forays into the intersection of AI systems with
copyright protection differ in their leanings, which shows how difficult
it can be to legislate comprehensively on AI issues. (Indeed, there are
already further proposed amendments to Japan's Copyright Act that
consider restricting the application of the 2018 amendment.) To date, we
haven't seen many court cases focused on generative models of text.
Perhaps the closest is a court case about computer programming language
code, namely
\href{https://blog.ericgoldman.org/archives/2023/06/how-can-ai-models-legally-obtain-training-data-doe-1-v-github-guest-blog-post.htm}{Doe
1 v. Github, Inc.}, which focuses on the fact that many public
repositories of code on the GitHub website, from which training data has
been drawn, come with a license that was stripped from the data during
training. Given that such court cases focus on training data, one
unanswered question is how such legal cases will affect companies'
openness about their models' training data in the future. As we
discussed, the more opaque the training data, the less hope we have of
understanding a model.

\hypertarget{how-can-you-contribute-to-a-healthy-ai-landscape}{%
\subsection{How can you contribute to a healthy AI
landscape?}\label{how-can-you-contribute-to-a-healthy-ai-landscape}}

There are a lot of important actions that help move us towards a future
where AI systems are developed in beneficial ways. We'll list a few
here.

\begin{itemize}
\tightlist
\item
  \textbf{If you're a student interested in AI systems:} you can become
  one of the people helping to decide how these models work. For anyone
  in this position, you'll find it useful to study computing, math,
  statistics, and also fields that reason about society. After all, the
  question of \emph{what} we build these systems to do deserves just as
  much attention as the question of \emph{how} we build these systems to
  do it.
\item
  \textbf{If you're an expert in something other than AI (e.g.,
  healthcare, a scientific or humanistic field):} the people building
  these models could really benefit from your expertise. Determining how
  to adapt AI systems to safely assist with problems faced by experts is
  not something computer scientists can do alone. To make these kinds of
  models useful for you and your field (and to avoid trying to solve
  problems that don't really need solving), model developers need your
  input and help. As more scientists and engineers enter the growing AI
  field, it should become easier to find people in your network who are
  working on the models. Engage with them!
\item
  \textbf{If you make decisions in a business sphere:} you can set a
  high bar for \emph{evaluating} possible AI-based systems in your
  company's workflow. There's considerable flashy language about some of
  these systems. By ignoring that and instead discussing with developers
  how a particular system was tested, how well that testing relates to
  your intended use case for it, and what's missing from those tests,
  you can help raise overall standards for evaluating AI.
\item
  \textbf{If you're a concerned consumer:} it's a huge help for you to
  assume a thoughtful, reflective distance about LMs and AI news. In
  recent months, there's been seemingly nonstop discussion of these
  topics, and there's sure to be a lot more coming. Our biggest goal for
  this document is that it will help to equip you with the knowledge you
  need to filter the hype and make sense of the substance.
\end{itemize}

\hypertarget{final-remarks}{%
\section{Final remarks}\label{final-remarks}}

Current language models are downright perplexing. By keeping in mind the
trends in the research communities that produced them, though, we gain a
sense of why these models behave as they do. Keeping in mind the primary
task that these models have been trained to accomplish, i.e., next word
prediction, also helps us to understand how they work. Many open
questions about these models remain, but we hope that we've provided
some helpful guidance on how to use and assess them. Though determining
how these technologies will continue to develop is difficult, there are
helpful actions that each of us can take to push that development in a
positive direction. By broadening the number and type of people involved
in decisions about model development and engaging in broader
conversations about the role of LMs and AI in society, we can all help
to shape AI systems into a positive force.

\hypertarget{acknowledgments}{%
\section*{Acknowledgments}\label{acknowledgments}}
\addcontentsline{toc}{section}{Acknowledgments}

The authors appreciate feedback from Sandy Kaplan, Lauren Bricker,
Nicole DeCario, and Cass Hodges at various stages of this project, which
was supported in part by NSF grant 2113530. All opinions and errors are
the authors' alone.

\hypertarget{glossary}{%
\section*{Glossary}\label{glossary}}
\addcontentsline{toc}{section}{Glossary}

\textbf{Algorithm}: A procedure that operates on a set of inputs in a
predefined, precisely specified way to produce a set of outputs.
Algorithms can be translated into computer programs. This document
references several different algorithms: (1) \textbf{stochastic gradient
descent}, which takes as input a (neural network) model
\textbf{architecture}, a dataset, and other settings and produces as
output a \textbf{model}; (2) a \textbf{model} itself, which takes as
input specified text and produces an output for the task the model was
trained to perform (for example, a \textbf{probability distribution}
over different kinds of attitudes being expressed for a sentiment
classification model, or a \textbf{probability distribution} over which
word comes next for a language model); (3) an algorithm for constructing
a language model's vocabulary (section \ref{datanuances}).

\textbf{Alignment} (of a \textbf{model} to human preferences): This term
can refer either to the degree to which a model reflects human
preferences, or to the process of adjusting a model to better reflect
human preferences. See section \ref{adaptingasproducts}.

\textbf{Architecture} (of a \textbf{model}): The template for arranging
a \textbf{model}'s \textbf{parameters} and specifying how those
parameters are jointly used (with an input) to produce the model's
output. Note that specifying the model architecture does \emph{not}
involve specifying the values of individual parameters, which are
defined later. (If you consider a model to be a ``black box'' with knobs
on its side that is given an input and produces an output, the model's
``architecture'' refers to the arrangement of knobs on/inside the box
\emph{without} including the particular values to which each knob is
set.)

\textbf{Artificial intelligence (AI)}: (1) Broadly describes several
fields or research communities that focus on improving machines' ability
to process complicated sources of information about the world (like
images or text) into predictions, analyses, or other human-useful
outputs. (2) Also refers in popular usage (but not this guide) to an
individual system (perhaps a \textbf{model}) built using techniques
developed in those fields (such as Deep Blue or ChatGPT).

\textbf{Bleu scores}: A fully automated way introduced by Papineni et
al. (\protect\hyperlink{ref-papineni-etal-2002-bleu}{2002}) to evaluate
the quality of a proposed translation of text into a target language. At
a high level, the Bleu score for a proposed translation of text (with
respect to a set of approved reference translations for that same text)
is calculated by looking at which fraction of small chunks (e.g.,
one-word chunks, two-word chunks, etc.) of the proposed translation
appear in at least one of the reference translations.

\textbf{Computer vision (CV)}: A subfield of computer science research
that advances the automated processing and production of information
from visual signals (images).

\textbf{Content safeguards}: A term commonly used within \textbf{NLP} to
refer to the strategies that are used to try to keep \textbf{language
models} from generating outputs that are offensive, harmful, dangerous,
etc. We give some examples of these strategies in section
\ref{safeguards}.

\textbf{Convergence}: A concept in \textbf{machine learning} that
explains when the \textbf{loss} between a \textbf{model}'s output and
expected output from \textbf{data} is less than some threshold. Model
convergence during training usually means that the model is no longer
improving, such as occurs at the end of \textbf{SGD}.

\textbf{Data}: The pairs of sample inputs and their desired outputs
associated with a \textbf{task}, used to train or evaluate a
\textbf{model}. For NLP, this is typically a massive collection of
either text that originates in digital form (e.g., text scraped from a
post published to an internet forum) or text converted into a digital
format (e.g., text extracted from a scanned handwritten document). It
may also include additional information describing the text, like
sentiment labels for a sentiment analysis dataset.

\textbf{Data-driven}: A description of a process indicating that it
determines actions based on analysis of massive data stores (in contrast
to having a person or multiple people make all of these decisions). For
example, a person deciding on the vocabulary for a \textbf{language
model} they're about to build could either (1) manually define a list of
all words or parts of words that the model's vocabulary would include
(not data-driven) or (2) collect text \textbf{data} and run a
data-driven \textbf{algorithm} (see section \ref{datanuances}) to
automatically produce a vocabulary based on that dataset for the
eventual model. \textbf{Machine learning} algorithms are, in general,
data-driven.

\textbf{Deep learning}: A term that describes \textbf{machine learning}
methods focused on training (\textbf{neural network}) \textbf{models}
with many \textbf{layers}.

\textbf{Depth} (of a \textbf{model}): Refers to the number of
\textbf{layers} a neural network architecture contains.

\textbf{Domain} (of \textbf{data}): A specific and intuitive (though not
formally defined) grouping of specific data. For example, an NLP
researcher might refer to ``the Wikipedia domain'' of text data, or
``the business email domain'' of text data. The term offers an expedient
way for researchers or practitioners to refer to data that generally has
some unifying characteristics or characteristics different from some
other data.

\textbf{Extrinsic evaluation} (of a \textbf{model}): An evaluation (of a
model) that evaluates whether using that model as part of a larger
system helps that system (and how much), or which considers factors
related to the model's eventual use in practice, etc.

\textbf{Finetuning} (of a \textbf{model} for a specific \textbf{task}):
Continued training of a model on a new dataset of choice that occurs
after original parameter values were trained on other tasks/datasets.
Use of the term ``finetuning'' indicates that the model about to be
finetuned has already been trained on some task/dataset.

\textbf{Function}: Broadly, a \textbf{mapping} of inputs to outputs. In
other words, a function takes as input any input matching a particular
description (like ``number'' or ``text'') and will give a (consistent)
answer for what that input's corresponding output should be. However,
everywhere we use the word ``function'' in this document (except in the
context of ``autocomplete functions''), we are referring more
specifically to functions that take in a set of numbers and produce
single-number outputs.

\textbf{Generative AI}: A subset of \textbf{artificial intelligence}
focused on \textbf{models} that learn to simulate (and can therefore
automatically produce/generate) complex forms of data, such as text or
images.

\textbf{Gradient} (of a function): A calculus concept. Given a
particular point in an \emph{n}-dimensional landscape, the gradient of a
\textbf{function} indicates the direction (and magnitude) of that
function's steepest ascent from that point. By considering the current
\textbf{parameters} of a \textbf{neural network} \textbf{model} as the
point in that \emph{n}-dimensional landscape, and taking the gradient of
a \textbf{loss function} with respect to those parameters, it is
possible to determine a very small change to each \textbf{parameter}
that \emph{increases} the loss function as much as locally possible.
This also indicates that the \emph{opposite} small change can
\emph{decrease} the loss function as much as locally possible, the goal
when running \textbf{SGD}.

\textbf{Hallucination} (by a \textbf{language model}): A term commonly
used to describe nonfactual or false statements in outputs produced by a
language model.

\textbf{Hardware}: The (physical) machines on which algorithms are run.
For contemporary NLP, these are typically GPUs (graphics processing
units), which were initially designed to render computer graphics
quickly but were later used to to do the same for the kinds of
matrix-based operations often performed by \textbf{neural networks}.

\textbf{Intrinsic evaluation} (of a \textbf{model}): An evaluation (of a
model) that evaluates that model on a specific \textbf{test set} ``in a
vacuum,'' that is, without considering how plugging that model into a
larger system would help that larger system.

\textbf{Label}: Some \textbf{tasks} have outputs that are a relatively
small set of fixed categories (unlike language modeling, where the
output is a \textbf{token} from some usually enormous vocabulary). In
cases where outputs are decided from that kind of small set,
\textbf{NLP} researchers typically refer to the correct output for a
particular input as that input's ``label''. For example, the set of
labels for an email spam-identification task would be ``spam'' or ``not
spam,'' and a \textbf{sentiment analysis} task might define its set of
possible labels to be ``positive,'' ``negative,'' or ``neutral.''

\textbf{Language model}: A \textbf{model} that takes text as input and
produces a \textbf{probability distribution} over which word in its
vocabulary might come next. See section \ref{lmtask}.

\textbf{Layer} (of a \textbf{neural-network}-based \textbf{model}): A
submodule with learnable parameters of a \textbf{neural network} that
takes as input a numerical representation of data and outputs a
numerical representation of data. Modern neural networks tend to be
\textbf{deep}, meaning that they ``stack'' many layers so that the
output from one layer is fed to another, whose output is then fed to
another, and so on.

\textbf{Loss function}: A mathematical \textbf{function} that takes in a
\textbf{model}'s proposed output given a particular input and compares
it to (at least) one reference output for what the output is
\emph{supposed} to be. Based on how similar the reference output is to
the model's proposed output, the loss function will return a single
number, called a ``loss.'' The higher the loss, the less similar the
model's proposed output is to the reference output.

\textbf{Machine learning (ML)}: An area of computer science focused on
\textbf{algorithms} that learn how to (approximately) solve a problem
from \textbf{data}, i.e., to use data to create other
\textbf{algorithms} (\textbf{models}) that are deployable on new,
previously unseen data.

\textbf{Mappings} (of input to output): A pairing of each (unique)
possible input to a (not necessarily unique) output, with the mapping
``translating'' any input it is given to its paired output.

\textbf{Model}: An \textbf{algorithm} for performing a particular
\textbf{task}. (NLP researchers typically refer to such an algorithm as
a model only if its corresponding task is sufficiently complicated to
lack any provably correct, computationally feasible way for a machine to
perform it. Hence, we apply \textbf{machine learning} to build a model
to approximate the task.) Though a model that performs a particular task
does not necessarily have to take the form of a \textbf{neural network}
(e.g., it could instead take the form of a list of human-written rules),
in practice, current \textbf{NLP} models almost all take the form of
neural networks.

\textbf{Natural language processing (NLP)}: A subfield of computer
science that advances the study and implementation of automated
processing and generation of information from text and, perhaps, other
language data like speech or video of signed languages.

\textbf{Neural network}: A category of \textbf{model}
\textbf{architecture} widely used in \textbf{machine learning} that is
subdifferentiable and contains many \textbf{parameters}, making it
well-suited to being trained using some variant of \textbf{stochastic
gradient descent}. Neural networks use a series of calculations
performed in sequence by densely connected \textbf{layers} (loosely
inspired by the human brain) to produce their output.

\textbf{(Numerical) optimization}: Can refer to (1) a family of
strategies for choosing the best values for a predetermined set of
\textbf{parameters,} given a particular quantity to minimize/maximize
which is calculated based on those parameters (and often some
\textbf{data} as well) or to (2) the field of research that studies
these strategies. In this document we refer exclusively to the first
definition.

\textbf{Overfitting}: When a \textbf{model} learns patterns that are
overly specific to its training \textbf{data} and that do not generalize
well to new data outside of that training set. This problem is typically
characterized by the model's very strong task performance on the
training data itself but far worse performance when given previously
unseen data.

\textbf{Parallel text}: A term used within \textbf{NLP} to refer to
pairs of text (usually pairs of sentences) in two languages that are
translations of each other. Parallel text is widely used for the
development of NLP \textbf{models} that perform the task (commonly
called ``machine translation'') of translating text from a specific
source language (e.g., Urdu) into a specific target language (e.g.,
Thai). Some pairs of languages have much more (digital) parallel text
available, and the difference in the quality of machine translation
systems across different language pairs reflects that disparity.

\textbf{Parameter} (in a neural network \textbf{model}): A single value
(model coefficient) that is part of the mathematical function that
neural networks define to perform their operations. If we consider a
\textbf{model} as being a black box that performs some \textbf{task,} a
parameter is a single one of that black box's knobs. ``Parameter'' can
refer either to the knob itself or the value the knob is set to,
depending on context.

\textbf{Perplexity}: A number from 1 to infinity that represents how
``surprised'' a \textbf{language model} generally is to see the actual
continuations of fragments of text. The lower the perplexity, the better
the language model can predict the actual continuations of those text
fragments in the evaluation data. Perplexity is an important
\textbf{intrinsic evaluation} for language models.

\textbf{Probability distribution}: A collection of numbers (not
necessarily unique) that are all at least 0 and add up to 1 (for
example, 0.2, 0.2, 0.1, and 0.5), each paired with some possible event;
the events are mutually exclusive. For one such event, its number is
interpreted as the chance that the event will occur. For example, if a
\textbf{language model} with a tiny vocabulary consisting of only
{[}apple, banana, orange{]} takes as input the sentence-in-progress
``banana banana banana banana'' and produces a probability distribution
over its vocabulary of 0.1 for ``apple,'' 0.6 for ``banana,'' and 0.3
for ``orange,'' this means that the model is predicting that the next
word to appear after the given sentence-in-progress has a 60\% chance of
being ``banana.''

\textbf{Prompt} (to a \textbf{language model}): The text provided by a
user to the language model, which the \textbf{model} then uses as its
context---i.e., as its initial basis for its next word prediction that
it performs over and over again to produce its output, word by word.

\textbf{Sentiment analysis}: A \textbf{task} in \textbf{NLP} that aims
to determine whether the overall sentiment of a piece of text skews
positive, negative, or in some versions of the task, neutral. For
example, suppose that a sentiment analysis \textbf{model} was given the
input ``Wow, that movie was amazing!'' The correct output for the model
given that input would be ``positive'' (or five stars, or 10/10, or
something similar if the labels were in the form of stars or integer
scores from 0 to 10 instead).

\textbf{Stochastic gradient descent (SGD)}: A process by which
parameters of a \textbf{model} are adjusted to minimize some specific
function (e.g., a \textbf{loss function}). SGD requires repeatedly
running varying batches of data through the model, whose output can then
be used to get a value from our (loss) function. For each batch, we then
use the \textbf{gradient} of that function to adjust the
\textbf{parameters} of our model to take a tiny descending step along
that gradient. This process is repeated until the loss function's
gradient flattens out and stops indicating a lower direction.

\textbf{Task}: A job we want a \textbf{model} to do. Tasks are usually
described abstractly---for example, sentiment analysis, question
answering, or machine translation---in a way that is not tied to any one
source of \textbf{data}. However, in practice, if a model is trained to
perform a particular task, the version of that task that the model
learns to perform will be heavily influenced by the specific training
data used. See section \ref{biases}.

\textbf{Test set} (or \textbf{test data}): A set of \textbf{data} unseen
by a \textbf{model} during its training, used to evaluate how well the
model works.

\textbf{Token}: The base unit of language into which an \textbf{NLP}
\textbf{model} splits any text input. For contemporary \textbf{language
models,} a token can be either a word or a piece of a word. A text input
passed to such a model will be split into its component words (in cases
where that word is part of the model's vocabulary) and word pieces (in
cases where that full word doesn't exist in the model's vocabulary, so
its component pieces are added to the sequence of tokens instead).

\textbf{Training set} (or \textbf{training data}): A set of
\textbf{data} used to train a \textbf{model} (in other words, to decide
that \textbf{model}'s parameter values). For a model that takes the form
of a \textbf{neural network}, the training set comprises the batches of
data used while running \textbf{stochastic gradient descent}.

\textbf{Transformer}: A kind of neural network \textbf{architecture}
introduced in 2017 that allows large \textbf{models} built using it to
train faster than earlier model architectures would have allowed, and on
more data (assuming access to certain relatively high-memory
\textbf{hardware}). They do this by using techniques (e.g.,
self-attention) beyond the scope of this work. See section
\ref{transformers}.

\hypertarget{references}{%
\section*{References}\label{references}}
\addcontentsline{toc}{section}{References}

\hypertarget{refs}{}
\begin{CSLReferences}{1}{0}
\leavevmode\vadjust pre{\hypertarget{ref-church-mercer-1993-introduction}{}}%
Church, Kenneth W., and Robert L. Mercer. 1993. {``Introduction to the
Special Issue on Computational Linguistics Using Large Corpora.''}
\emph{Computational Linguistics} 19 (1): 1--24.
\url{https://aclanthology.org/J93-1001}.

\leavevmode\vadjust pre{\hypertarget{ref-dodge-etal-2021-documenting}{}}%
Dodge, Jesse, Maarten Sap, Ana Marasović, William Agnew, Gabriel
Ilharco, Dirk Groeneveld, Margaret Mitchell, and Matt Gardner. 2021.
{``Documenting Large Webtext Corpora: A Case Study on the Colossal Clean
Crawled Corpus.''} In \emph{Proceedings of the 2021 Conference on
Empirical Methods in Natural Language Processing}, 1286--1305. Online;
Punta Cana, Dominican Republic: Association for Computational
Linguistics. \url{https://doi.org/10.18653/v1/2021.emnlp-main.98}.

\leavevmode\vadjust pre{\hypertarget{ref-gopnik}{}}%
Gopnik, Alison. 2023. {``{Large Language Models as Cultural
Technologies}.''} Presented at the 61st {A}nnual {M}eeting of the
{A}ssociation for {C}omputational {L}inguistics.

\leavevmode\vadjust pre{\hypertarget{ref-gururangan-etal-2022-whose}{}}%
Gururangan, Suchin, Dallas Card, Sarah Dreier, Emily Gade, Leroy Wang,
Zeyu Wang, Luke Zettlemoyer, and Noah A. Smith. 2022. {``Whose Language
Counts as High Quality? Measuring Language Ideologies in Text Data
Selection.''} In \emph{Proceedings of the 2022 Conference on Empirical
Methods in Natural Language Processing}, 2562--80. Abu Dhabi, United
Arab Emirates: Association for Computational Linguistics.
\url{https://aclanthology.org/2022.emnlp-main.165}.

\leavevmode\vadjust pre{\hypertarget{ref-chinchilla}{}}%
Hoffmann, Jordan, Sebastian Borgeaud, Arthur Mensch, Elena Buchatskaya,
Trevor Cai, Eliza Rutherford, Diego de Las Casas, et al. 2022. {``An
Empirical Analysis of Compute-Optimal Large Language Model Training.''}
In \emph{Advances in {N}eural {I}nformation {P}rocessing {S}ystems},
edited by S. Koyejo, S. Mohamed, A. Agarwal, D. Belgrave, K. Cho, and A.
Oh, 35:30016--30. Curran Associates, Inc.
\url{https://proceedings.neurips.cc/paper_files/paper/2022/file/c1e2faff6f588870935f114ebe04a3e5-Paper-Conference.pdf}.

\leavevmode\vadjust pre{\hypertarget{ref-alondranelson}{}}%
Nelson, Alondra. 2023. {``Thick Alignment.''} Presented at the 2023
{ACM} {C}onference on {F}airness, {A}ccountability, and {T}ransparency
({ACM FA}cc{T}). \url{https://youtu.be/Sq_XwqVTqvQ?t=957}.

\leavevmode\vadjust pre{\hypertarget{ref-nishi2019privatizing}{}}%
Nishi, Andrea. 2019. {``Privatizing Sentencing: A Delegation Framework
for Recidivism Risk Assessment.''} \emph{Columbia Law Review} 119 (6):
1671--1710.
\url{https://columbialawreview.org/content/privatizing-sentencing-a-delegation-framework-for-recidivism-risk-assessment/}.

\leavevmode\vadjust pre{\hypertarget{ref-papineni-etal-2002-bleu}{}}%
Papineni, Kishore, Salim Roukos, Todd Ward, and Wei-Jing Zhu. 2002.
{``{B}leu: A Method for Automatic Evaluation of Machine Translation.''}
In \emph{Proceedings of the 40th Annual Meeting of the Association for
Computational Linguistics}, 311--18. Philadelphia, Pennsylvania, USA:
Association for Computational Linguistics.
\url{https://doi.org/10.3115/1073083.1073135}.

\leavevmode\vadjust pre{\hypertarget{ref-elmo}{}}%
Peters, Matthew E., Mark Neumann, Mohit Iyyer, Matt Gardner, Christopher
Clark, Kenton Lee, and Luke Zettlemoyer. 2018. {``Deep Contextualized
Word Representations.''} In \emph{Proceedings of the 2018 Conference of
the North {A}merican Chapter of the Association for Computational
Linguistics: Human Language Technologies, Volume 1 (Long Papers)},
2227--37. New Orleans, Louisiana: Association for Computational
Linguistics. \url{https://doi.org/10.18653/v1/N18-1202}.

\leavevmode\vadjust pre{\hypertarget{ref-Prates2019}{}}%
Prates, Marcelo O. R., Pedro H. Avelar, and Luís C. Lamb. 2019.
{``Assessing Gender Bias in Machine Translation: A Case Study with
Google Translate.''} \emph{Neural Computing and Applications} 32 (10):
6363--81. \url{https://doi.org/10.1007/s00521-019-04144-6}.

\leavevmode\vadjust pre{\hypertarget{ref-raghavan-hiring}{}}%
Raghavan, Manish, Solon Barocas, Jon Kleinberg, and Karen Levy. 2020.
{``Mitigating Bias in Algorithmic Hiring: Evaluating Claims and
Practices.''} In \emph{Proceedings of the 2020 Conference on Fairness,
Accountability, and Transparency}, 469--81. FAT* '20. New York, NY, USA:
Association for Computing Machinery.
\url{https://doi.org/10.1145/3351095.3372828}.

\leavevmode\vadjust pre{\hypertarget{ref-sap-etal-2019-risk}{}}%
Sap, Maarten, Dallas Card, Saadia Gabriel, Yejin Choi, and Noah A.
Smith. 2019. {``The Risk of Racial Bias in Hate Speech Detection.''} In
\emph{Proceedings of the 57th Annual Meeting of the Association for
Computational Linguistics}, 1668--78. Florence, Italy: Association for
Computational Linguistics. \url{https://doi.org/10.18653/v1/P19-1163}.

\leavevmode\vadjust pre{\hypertarget{ref-sennrich-etal-2016-neural}{}}%
Sennrich, Rico, Barry Haddow, and Alexandra Birch. 2016. {``Neural
Machine Translation of Rare Words with Subword Units.''} In
\emph{Proceedings of the 54th Annual Meeting of the Association for
Computational Linguistics (Volume 1: Long Papers)}, 1715--25. Berlin,
Germany: Association for Computational Linguistics.
\url{https://doi.org/10.18653/v1/P16-1162}.

\leavevmode\vadjust pre{\hypertarget{ref-shannonsgame}{}}%
Shannon, C. E. 1951. {``Prediction and Entropy of Printed English.''}
\emph{The Bell System Technical Journal} 30 (1): 50--64.
\url{https://doi.org/10.1002/j.1538-7305.1951.tb01366.x}.

\leavevmode\vadjust pre{\hypertarget{ref-shneiderman-97}{}}%
Shneiderman, Ben, and Pattie Maes. 1997. {``Direct Manipulation Vs.
Interface Agents.''} \emph{Interactions} 4 (6): 42--61.
\url{https://doi.org/10.1145/267505.267514}.

\leavevmode\vadjust pre{\hypertarget{ref-Vaswani}{}}%
Vaswani, Ashish, Noam Shazeer, Niki Parmar, Jakob Uszkoreit, Llion
Jones, Aidan N Gomez, Łukasz Kaiser, and Illia Polosukhin. 2017.
{``Attention Is All You Need.''} In \emph{Advances in Neural Information
Processing Systems}, edited by I. Guyon, U. Von Luxburg, S. Bengio, H.
Wallach, R. Fergus, S. Vishwanathan, and R. Garnett. Vol. 30. Curran
Associates, Inc.
\url{https://proceedings.neurips.cc/paper_files/paper/2017/file/3f5ee243547dee91fbd053c1c4a845aa-Paper.pdf}.

\leavevmode\vadjust pre{\hypertarget{ref-norbertwiener}{}}%
Wiener, Norbert. 1960. {``Some Moral and Technical Consequences of
Automation.''} \emph{Science} 131 (3410): 1355--58.
\url{http://www.jstor.org/stable/1705998}.

\end{CSLReferences}

\section*{Appendix}  
\addcontentsline{toc}{section}{Appendix}

\subsection*{Loss functions and gradient descent, a bit more formally}
\addcontentsline{toc}{subsection}{Loss functions and gradient descent, a bit more formally}

The first important property for a loss function is that \textit{it takes into account all the potential good and bad things about outputs} when deducting points. The more dissimilar our model's output given a particular input is from that input's correct output, the higher the loss function should be. The second important property is that \textit{we must be able to deduce, fully automatically and in parallel for all parameters, what adjustments would make the loss function decrease}. You may recall from a course on calculus that questions like ``How does a small change to an input to a function affect the function's output?'' are related to the concept of differentiation. In sum, we need the loss function to be differentiable with respect to the parameters. (This may be a bit confusing because in calculus, we think about differentiating a function with respect to its inputs. In a mathematical sense, the input is only part of the input to the mathematical function encoded by a neural network; the parameters are also part of its input.) If the loss function has this property, then we can use differentiation to automatically calculate a small change for each parameter that should decrease the loss on a given example.

These two properties\textemdash faithfulness to the desired evaluation and differentiability with respect to parameters\textemdash conflict because most evaluation scores aren't differentiable. Bleu scores for translation and error rates for sentiment analysis are stepwise functions (``piecewise constant'' in mathematical terms): changing the parameters a tiny bit usually won't affect these evaluation scores; when it does, it could be a dramatic change. Human judgments also are not differentiable with respect to parameters.

Once we know a differentiable loss function, and with a few additional assumptions, we quickly arrive at the algorithm for stochastic gradient descent (SGD), for setting system parameters. To describe its steps a bit more formally than we did in section~\ref{minimization}:

\begin{enumerate}
\item{Initialize the parameters randomly.}
\item{Take a random sample of the training data (typically 100 to 1000 demonstrations); run each input through the system and calculate the loss and its first derivative with respect to every parameter. (When first derivatives are stacked into a vector, it's called the \textbf{gradient}.) Keep a running total of the sum of loss values and a running total of the sum of gradients.}
\item{For each parameter, change its value proportional to the corresponding value in the gradient vector. (If the gradient is zero, don't change that parameter.)}
\item{Go to step 2 if the loss is converging.}
\end{enumerate}

\subsection*{Word error rate, more formally}
\addcontentsline{toc}{subsection}{Word error rate, more formally}

Given some test data (some text the language model wasn't trained on), we can calculate the error rate as follows.  Let the words in the test data be denoted by $w_1$, $w_2$, ... , $w_N$.
\begin{enumerate}
\item{Set $m = 0$; this is the count of mistakes.}
\item{For every word $w_i$ in the test data ($i$ is its position):}
     \begin{enumerate}
	\item[1.]{Feed $w_i$'s preceding context, which after the first few words will be the sequence $w_1$, $w_2$, ... , $w_{i - 1}$, into the language model as input.}
	\item[2.]{Let the language model predict the next word; call its prediction $w_{\textrm{pred}}$.}
	\item[3.]{If $w_{\textrm{pred}}$ is anything other than $w_i$, the language model made an incorrect prediction, so add 1 to $m$.}
     \end{enumerate}
\item{The error rate is $m / N$.}
\end{enumerate}

\subsection*{Perplexity, more formally}
\addcontentsline{toc}{subsection}{Perplexity, more formally}

Section~\ref{perplexity} describes underlying properties of \textit{how} LMs make ``decisions'' about next words. Here, to prepare for a deeper dive into perplexity, we summarize and build on those properties:

\begin{itemize}
\item{Based on the context of preceding words, a calculation is made by the neural network that assigns a \textit{probability} to every word in the vocabulary, that is, every possible choice of what word could come next. These probabilities must always sum to one (that's part of the definition of a probability distribution), and we also impose a ``\textbf{no zeros}'' rule: the probability of every vocabulary word must always be at least slightly positive.}
\item{To predict the next word, the model can either (a) choose the one with the highest probability (as assumed in the error rate calculation above) or (b) simulate a draw from the probability distribution, choosing a word at random such that each word's chance of being drawn is given by its probability. To illustrate, imagine a pub trivia team where individual members have different past success rates of being correct. Approach (a) would correspond to the team always submitting the answer proposed by the trivia-whiz team member whose suggested answers had most often been correct before. Approach (b) would correspond to randomly picking who should answer, with the trivia whiz's answer being most likely to be chosen, the second-best team member's answer next most likely, then the third-best team member's answer, and so on. Note that the most likely outcome from (b) is the same as the outcome from (a), but (b) will sometimes lead to another, lower-probability word.}
\end{itemize}

Whether (a), (b), or some other approach is used when an LM is deployed is an important design decision. In keeping with our earlier rejection of error rate, researchers try to avoid evaluating LMs in a way that makes unnecessary commitments to its eventual use.\footnote{The technical term for our desired evaluation is ``intrinsic'' evaluation, meaning that we want a measure of the intrinsic quality of a model, not its performance in some extrinsic setting.} Option (b) is interesting because it suggests a workaround to the pitfalls of simply counting mistakes discussed in section~\ref{perplexity}.

In the preceding appendix subsection's error rate calculation procedure, we could apply option (b) in step 2.2.  Suppose we do this not once, but many times for each context/word pair and average the error rate across these random draws. With enough draws, this approach would provide meaningful error rates because we'd expect to get each word right \textit{some} of the time (no zeros rule). In practice, rather than actually carrying out the random draws, we instead use the LM's probabilities \textit{directly} to assign a score for every word in the test data. The results of this approach are that:

\begin{itemize}
\item{If the language model gave probability 1 to the correct next word, the score for that word would be 1. This can't happen exactly because the probabilities of all the wrong words have to exceed zero (no zeros rule).  But we can get arbitrarily close in principle if the probabilities of all the wrong words get infinitesimally small.}
\item{If the LM gave probability 0 to the correct next word, the score for that word would be 0. But this can't happen either because of the no zeros rule.}
\item{In general, the greater the probability the LM assigns to the correct next word, even if it's not the most probable word, the higher the score.}
\end{itemize}

Because of the no zeros rule, the per-word probability scores are always somewhere between 0 and 1.

Given the test data, we can calculate the LM probability for every word given its preceding context. If we took a simple average of these probability scores and subtracted that from 1, we would get something like an error rate (technically, an ``expected'' error rate under prediction method (b)).  What is done in practice is similar in spirit but slightly different: we take the geometric average of the inverses of these probability scores, a value known as (test data) perplexity. The reasons are partly practical (tiny numbers can lead to a problem in numerical calculations, called underflow), partly theoretical, and partly historical.  For completeness, here's the procedure:

\begin{enumerate}
\item{Set $m = 0$.  (This quantity is no longer a running tally of mistakes.)}
\item{For every word $w_i$ in the test data ($i$ is its position):}
    \begin{enumerate}
	\item[1.]{Feed $w_i$'s preceding context, which after the first few words will be the sequence $w_1$, $w_2$, ... , $w_{i - 1}$, into the language model as input.}
	\item[2.]{Let $p$ be the probability that the language model assigns to $w_i$ (the correct next word).}
	\item[3.]{Add $-\log(p)$ to $m$.}
    \end{enumerate}
\item{The perplexity is $\exp(m / N)$.}
\end{enumerate}

Though it's probably not very intuitive from the preceding procedure, perplexity does have some nice intuitive properties:

\begin{itemize}
\item{If our model perfectly predicted every word in the test data with probability 1, we would get a perplexity of 1.\footnote{To see this, note that $-\log(1) = 0$, so $m$ stays 0 throughout step 2.  Note that $\exp(0 / N) = \exp(0) = 1$.} This can't happen because (1) there is some fundamental amount of uncertainty in fresh, unseen text data, and (2) some probability mass is reserved for every wrong word, too (no zeros rule). If perplexity comes very close to 1, the cardinal rule that test data must not be used for anything other than the final test, like training, should be carefully verified.}
\item{If our model ever assigned a probability of 0 to some word in the test data, perplexity would go to infinity.\footnote{To see this, note that $\log(0)$ tends toward infinity.} This won't happen because of the no zeros rule.}
\item{Lower perplexity is better.}
\item{The perplexity can be interpreted as an average ``branch factor''; in a typical next word prediction instance, how many vocabulary words are ``effectively'' being considered?}
\end{itemize}

\end{document}